\documentclass[10pt,journal,compsoc]{IEEEtran}
\usepackage{microtype}
\usepackage{graphicx}
\usepackage{subcaption}
\usepackage{booktabs} 
\usepackage{hyperref}

\newcommand{\eg}{\emph{e.g.,} }
\newcommand{\ie}{\emph{i.e.,} }
\newcommand{\jiasi}[1]{\textcolor{red}{[Jiasi: #1]}}
\newcommand{\xuechen}[1]{\textcolor{brown}{[Xuechen: #1]}}
\newcommand{\green}[1]{\textcolor{darkgreen}{#1}}
\newcommand{\sh}{\hat{\vct{s}}}

\usepackage{wrapfig}
\usepackage{hyperref,graphicx,amsmath,amssymb,bm,breakurl,epsfig,epsf,color,mathbbol,fmtcount,semtrans,multirow,comment,boldline,movie15,pgfplots}
\usepackage{tikz}
\usetikzlibrary{pgfplots.groupplots}
\usepackage[utf8]{inputenc} 
\usepackage[T1]{fontenc}    
\usepackage{url}            
\usepackage{booktabs}       
\usepackage{amsfonts}       
\usepackage{nicefrac}       
\usepackage{microtype}      
\usepackage{mathtools}
\definecolor{darkred}{RGB}{150,0,0}
\definecolor{darkgreen}{RGB}{0,150,0}
\definecolor{darkblue}{RGB}{0,0,200}
\definecolor{clor}{RGB}{125,0,125}
\hypersetup{colorlinks=true, linkcolor=darkred, citecolor=darkgreen, urlcolor=darkblue}

\numberwithin{equation}{section}

\newcommand{\clr}[1]{\textcolor{black}{#1}}
\newcommand{\cln}[1]{\textcolor{red}{}}
\newcommand{\inc}[1]{\textcolor{black}{#1}}

\def \endprf{\hfill {\vrule height6pt width6pt depth0pt}\medskip}

\newcommand{\tsn}[1]{{\left\vert\kern-0.25ex\left\vert\kern-0.25ex\left\vert #1 
    \right\vert\kern-0.25ex\right\vert\kern-0.25ex\right\vert}}

\newcommand{\sml}[1]{{\small{#1}}}
\newcommand{\sm}{s^m}
\newcommand{\ga}{f_c}

\newcommand{\gp}{\hat{g}}
\newcommand{\hsm}{\hat{s}^m}

\newcommand{\beq}{\begin{equation}}
\newcommand{\ba}{\begin{align}}
\newcommand{\ea}{\end{align}}

\newcommand{\eeq}{\end{equation}}

\newcommand{\Dc}{{\cal{D}}}

\newcommand{\Gc}{{\cal{G}}}

\newcommand{\onebb}{{\mathbf{1}}}

\newcommand{\Sc}{\mathcal{S}}

\newcommand{\s}{\vct{s}}

\newcommand{\g}{{\vct{g}}}

\newcommand{\Xc}{\mathcal{X}}
\newcommand{\Yc}{\mathcal{Y}}

\newcommand{\yh}{\hat{\y}}

\newcommand{\gh}{\hat{\g}}
\newcommand{\hg}{\hat{g}}

\newcommand{\x}{\vct{x}}

\newcommand{\y}{\vct{y}}

\newcommand{\bgl}{{~\big |~}}

\definecolor{emmanuel}{RGB}{255,127,0}

\newcommand{\R}{\mathbb{R}}
\newcommand{\Pro}{\mathbb{P}}

\newcommand{\E}{\operatorname{\mathbb{E}}}

\newcommand{\vct}[1]{\bm{#1}}

\newcommand{\red}[1]{\textcolor{black}{#1}}

\usepackage{changepage}
\usepackage{makecell}
\ifCLASSOPTIONcompsoc
  \usepackage[nocompress]{cite}
\else
  \usepackage{cite}
\fi
\ifCLASSINFOpdf
\else
\fi
\begin{document}
%
\title{Post-hoc Models for Performance Estimation of Machine Learning Inference}
%
%
%
%

\author{Xuechen Zhang
        \quad\quad\quad\quad
        Samet Oymak
        \quad\quad\quad\quad
        Jiasi Chen
\IEEEcompsocitemizethanks{\IEEEcompsocthanksitem X.~Zhang and S.~Oymak are with the Department of Electrical and Computer Engineering, University of California, Riverside. \protect\\
J.~Chen is with the Department of Computer Science and Engineering, University of California, Riverside.\protect\\
E-mails: xzhan394@ucr.edu, oymak@ece.ucr.edu, jiasi@cs.ucr.edu.}
}

%
%

\markboth{Journal of \LaTeX\ Class Files,~Vol.~14, No.~8, August~2015}%
{Shell \MakeLowercase{\textit{et al.}}: Bare Demo of IEEEtran.cls for Computer Society Journals}
\markboth{}{}
%



\IEEEtitleabstractindextext{%
\begin{abstract}

Estimating how well a machine learning model performs during inference is critical in a variety of scenarios (for example, to quantify uncertainty, or to choose from a library of available models). However, the standard accuracy estimate of softmax confidence is not versatile and cannot reliably predict different performance metrics (e.g., F1-score, recall) or the performance in different application scenarios or input domains. In this work, we systematically generalize performance estimation to a diverse set of metrics and scenarios and discuss generalized notions of uncertainty calibration. We propose the use of post-hoc models to accomplish this goal and investigate design parameters, including the model type, feature engineering, and performance metric, to achieve the best estimation quality. Emphasis is given to object detection problems and, unlike prior work, our approach enables the estimation of per-image metrics such as recall and F1-score. Through extensive experiments with computer vision models and datasets in three use cases -- mobile edge offloading, model selection, and dataset shift -- we find that proposed post-hoc models consistently outperform the standard calibrated confidence baselines. To the best of our knowledge, this is the first work to develop a unified framework to address different performance estimation problems for machine learning inference.
\end{abstract}
}

\maketitle

\IEEEdisplaynontitleabstractindextext

%
\IEEEpeerreviewmaketitle

\IEEEraisesectionheading{\section{Introduction}\label{sec:introduction}}

%
%
%
%


Machine learning inference pipelines typically do not have any way of knowing how well they are doing \inc{during runtime, beyond the softmax probability score of the model}.
However, an estimate of the current inference performance \inc{for different applications} can be very useful for a variety of purposes. \inc{For instance, in natural language processing, we might want to estimate the quality of a neural translation in terms of BLEU score. In object detection, we may want to estimate the F1-score and mean Average Precision (mAP), which provide critical summaries of the output quality but are not accessible during inference.} 
In edge computing~\cite{ran2018deepdecision,zhang2018awstream}, resource allocation decisions are made based on the estimated inference performance on the test set. 
Critically, all such decisions assume that the performance on the test domain is known, or rely on either similar distributional characteristics for the test and training domains. In this work, we seek to address this key stumbling block by asking
\begin{adjustwidth}{0.7cm}{}
\textbf{Q:}~Can we estimate the performance of black-box models across different metrics, domains, \& applications?
\end{adjustwidth}
The key hypothesis is that it is possible to make accurate, per-example predictions of how well a DNN will perform, and utilize these predictions to improve practical use cases.
To make these predictions quickly, we focus our design space on lightweight post-hoc models that operate on the outputs of the black-box DNN model that is performing the main inference.
\clr{While there has been work on specific instances of performance estimation, such as predicting resource consumption or image segmentation quality~\cite{robinson2018real,lu2017modeling}, to the best of our knowledge, this is the first work to provide a general framework for performance prediction of DNNs in a post-hoc fashion.} 

Related work has used calibrated confidence as an estimate of inference performance~\cite{platt1999probabilistic,naeini,zadrozny2001obtaining,guo2017calibration,naeini2015obtaining,zadrozny2002transforming,hendrycks2019using,Kuppers_2020_CVPR_Workshops}, but our approach is more general and can accommodate a richer set of input features and output performance metrics. For instance, our proposed framework can estimate the performance gap across different models, in addition to the performance of a single model.
Importantly, our evaluations show that our approach handily outperforms such confidence-based approaches. \red{Additionally, to assess estimation quality, prior works are mostly restricted to variations of Expected Calibration Error (ECE), which as we discuss in Section \ref{sec:setup}, may not be suitable for certain applications.}
Works specific to object detection~\cite{Kuppers_2020_CVPR_Workshops,schwaiger2021black,schubert2020metadetect,rahman2020performance} focus on estimating the uncertainty of the location or scale of a given detected object, but cannot evaluate the image as a whole, such as how many objects were missed (false negatives) which is needed to predict per-image metrics F1 score or recall. \red{Perhaps more importantly, these works provide point solutions and do not address the growing need for general-purpose methodologies that can be effortlessly adapted to new problems of interest.}

When designing our general framework to work for different use cases, a variety of challenges arise. How to define a general framework that can adapt to different use cases? What is the right choice of input features and design for the post-hoc model? 
Can the post-hoc model work well even if there are few samples? These questions become particularly challenging for applications beyond standard image classification.
For instance, in object detection, models have very high-dimensional outputs (capturing the logits and locations of multiple objects), and the output dimension is variable (depending on the number of detected objects). 
To overcome these challenges, we systematically explore different post-hoc model designs and input feature choices, to develop a post-hoc model that works well even with few number of training samples.

Overall, this work addresses both high-level and application-specific challenges through the following contributions.
\begin{itemize}
\item \textbf{General Framework for Post-Hoc Model Design:} We formulate the general problem of post-hoc model design, which is flexible enough to accomodate a variety of performance metrics and input features based on the desired machine learning inference pipeline. This also leads to a natural notion of \emph{calibration error} for a general class of performance metrics.
We describe how our framework applies to three practical use cases, simply by modifying certain definitions in the framework;
these use cases include (a) choosing between multiple machine learning inference models, (b) deciding whether to offload machine learning inference from a mobile device to an edge server, and (c) calibrating models after dataset shift.

\item \textbf{\clr{Applications, Experiments, \& Insights}:} We perform extensive numerical experiments of object detection and image classification to show the efficacy of our approach, compared to the baseline of using calibrated confidence as an estimate of inference performance. We show that metrics such as F1 score, precision, and recall can be accurately predicted by our post-hoc model and outperforms a calibrated confidence baseline. \red{For instance, in COCO dataset while confidence-based approach achieves 2.67\% Expected Calibration Error (ECE) we achieve 1.65\% ECE for F1 score prediction.} 
We show that our post-hoc model can accurately predict the inference performance of the three use cases described above, on different datasets (COCO, VOC) and models (SSD, MobileNets, YOLO, etc.).
 \inc{A key finding is that performing intelligent feature selection and reducing the dimensionality of the black-box model outputs (which are the inputs to the post-hoc model) can greatly reduce sample complexity and enhance the performance estimates.} \red{Finally, besides commonly used ECE, we propose \emph{Spearman's rank correlation} as an alternative metric to assess calibration performance and discuss its potential benefits over ECE.}
\end{itemize}

The remainder of this paper is organized as follows.
In Section \ref{sec:related}, we discuss related work.
In Section \ref{sec:framework}, we describe our general framework, followed by numerical results in Section \ref{sec:numerics}.
Finally, we conclude in Section \ref{sec:conclusions}.

\section{Related Work}
\label{sec:related}

\textbf{Confidence calibration:} Confidence is one possible metric of inference performance, and there are a wealth of confidence calibration approaches in the literature, such as Platt/Temperature scaling \cite{platt1999probabilistic}, Histogram binning \cite{zadrozny2001obtaining}, Bayesian Binning into Quantiles (BBQ) \cite{naeini2015obtaining}, Isotonic regression \cite{zadrozny2002transforming}, and Platt scaling extensions \cite{guo2017calibration}. 
Several works \cite{lakshminarayanan2017simple,de2018clinically,thulasidasan2019mixup,kumar2019verified,zhao2020role,hendrycks2019using,naeini,kumar2018trainable} consider either algorithmic improvements or application specific challenges associated to uncertainty quantification. \cite{snoek2019can} studies model uncertainty under dataset shift and provides empirical comparison of different calibration techniques. 
This work studies performance estimation metrics beyond confidence that are specific to the application, such as F1-score for object detection, which may be more interpretable by practitioners (\eg ``what is the predicted F1 score of this test image?'' is more interpretable than ``what is the predicted confidence of this test image?'').
\textbf{Confidence of object detection:} Several works have examined confidence calibration for object detection specifically, which is also the application domain that this work focuses on. 
  \cite{Kuppers_2020_CVPR_Workshops} presents a framework to measure and calibrate biased (or miscalibrated) confidence estimates of the model output. This approach results in calibrated confidence estimates of the object location and box scale. \cite{schwaiger2021black} \clr{additionally considers the impact of post-processing methods, such as non-maximum suppression, on confidence calibration.}
  However, these works focus on performance estimates per object, whereas this work studies more general per image performance estimation metrics.
  These works are only able to calibrate confidence for objects that are detected in the image, and miss on those objects that were false negatives.
In contrast, our approach can calibrate metrics that incoporate false negatives (such as F1 score or recall).

  \textbf{Other performance metrics:} Prediction of other performance metrics (\eg segmentation quality, intersection over union) have also been studied in the literature. \cite{schubert2020metadetect} proposes meta-regression to predict the intersection over union (IoU), and also classifies true and false positives. \cite{rahman2020performance} predicts when the per-frame mAP drops below a critical threshold. 
  \cite{rottmann2020prediction,robinson2018real} predict segmentation quality.
  \cite{ovadia2019can} evaluated prediction uncertainty after dataset shift.
  Our work provides a more general framework that can encompass these disparate performance metrics.

  The area of image quality assessment (IQA)~\cite{wang2006modern} seeks to predict the perceptual quality of images, with recent work using DNNs to perform the prediction~\cite{yang2019survey}.
  IQA has some similarity to this work in that IQA also predicts various (perceptual) metrics, and the original source data is also images; however, the main difference is that we seek to predict one step further down the computational pipeline -- not the quality of an image itself, but the quality of a prediction generated from that image.

\section{Performance Estimation Framework}
\label{sec:framework}

\subsection{Post-hoc predictions of scores}
We first describe the problem setup and our general framework for designing our post-hoc model.
Let $f:\Xc\rightarrow\R^k$ be the machine learning model of interest for which we want to estimate the performance; for example, a deep neural network that maps input space into a $k$ dimensional output. Let $(\x,\y)\sim\Dc$ be the distribution of our dataset. Given an input $\x\in\Xc$, the model outputs a prediction $\yh=f(\x)$. For image recognition, $\yh$ is a probability distribution over the classes. However, in more practical scenarios, $\yh$ can contain more complex features. For instance, for object detection, $\yh$ contains the likelihoods of multiple objects, as well as their locations and sizes in the input image. During inference time, it is desirable to know the performance of the classification. This performance can be assessed through a performance metric $m(\y,\yh)$.
For example, $m$ could be defined as the F1 score, recall, mean Average Precision (mAP), \inc{BLEU score} etc.
We also define a score function $\sm(\cdot)$ which is a processed version of the raw performance metric $m$ via
\[
\sm(\cdot)=s\circ m(\cdot).
\]
In simple use cases, we set the processing function as $s \gets \text{\emph{identity}}$, and in more complex use cases, $\sm$ is allowed to be a sophisticated function of $m$.

\begin{figure}
  \centering
    \includegraphics[width=0.5\textwidth]{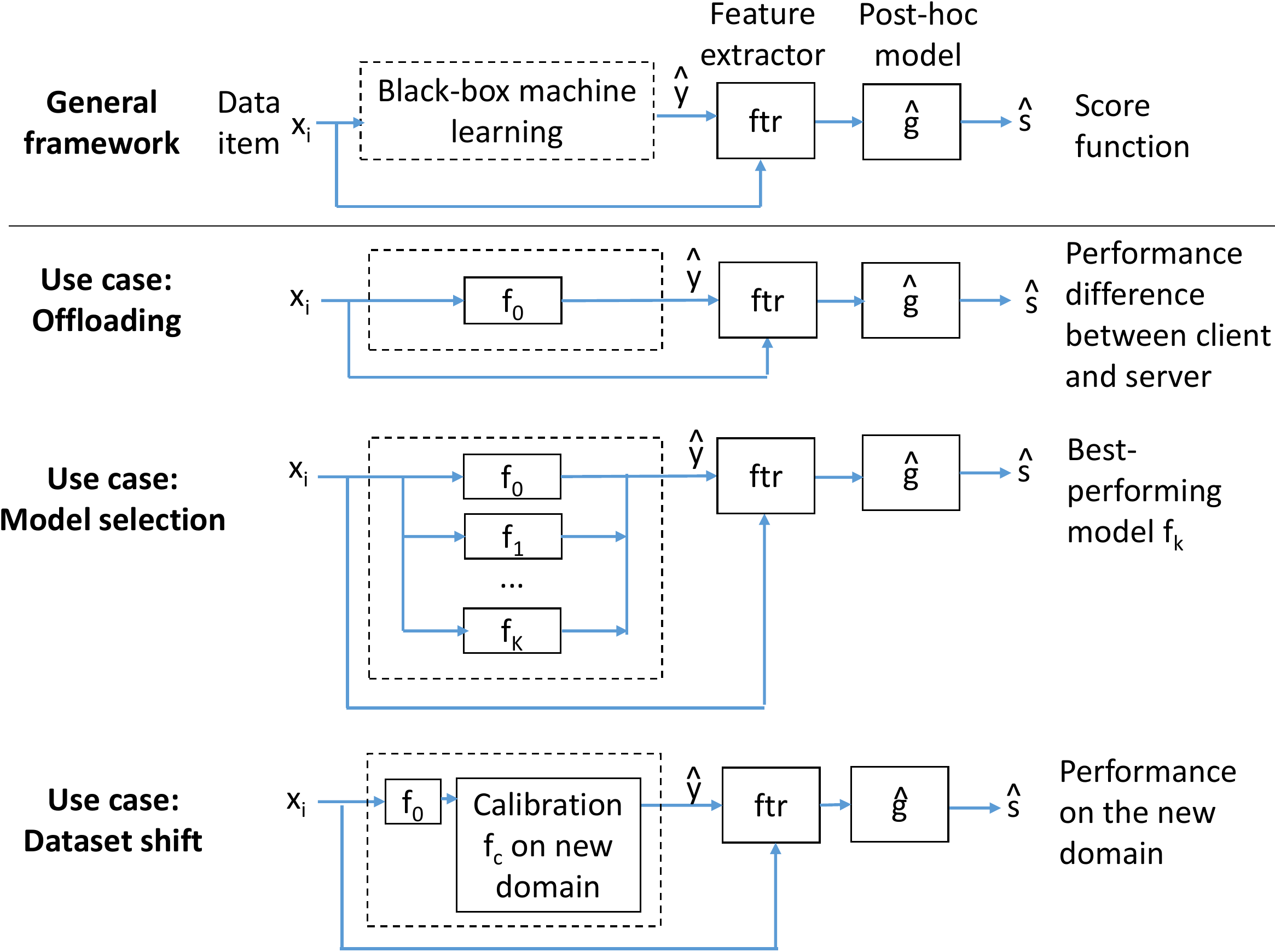}
    \caption{Overview of our performance estimation framework and its application to three use cases. 
The input data $x_i$ is fed into a black-box machine learning model. The output prediction $\hat{y}$ is fed into a feature extractor (ftr), and from there into the post-hoc model $\hat{g}$. The final output is the score function $\hat{s}$. The black-box machine learning model, features, and performance estimate are defined based on the individual use cases (see Table~\ref{table:summary}). \red{Jiasi to update figure} }
    \label{fig:diagram}
\end{figure}

Our goal is to build a post-hoc model $g$ that estimates the performance $\hat{\s}$ of the model $f$ during inference. Let $\ell(\s,\sh)$ be a loss function that takes the true performance $\s$ and the estimated performance $\sh$ as inputs. Let $\Sc$ be a training dataset for post-hoc modeling. 
Typically $\Sc=\{(\x_i,y_i)\}_{i=1}^n$ is a secondary training dataset of size $n$ independent of the one used to build $f$. \inc{When training multiple models -- as in the example of dataset shift use case (Section \ref{sec:domain_adaptation}) -- we will split $\Sc$ into disjoint sets (\eg train-validation).} We use $\Sc$ to fit the model $g$ via
\begin{align}
\hat{g}=\arg\min_{g\in \Gc}\frac{1}{n}\sum_{i=1}^n\ell(\sm(\y_i,f(\x_i)),g\circ f(\x_i)).\label{score pred}
\end{align}

An overview of our approach is shown in the top row of Fig.~\ref{fig:diagram}, with details on each of the use cases given in Table~\ref{table:summary}.

\begin{table}[]
\small
\begin{tabular}{|p{2cm}|p{2.5cm}|p{2.5cm}|}
\hline
                                  Use case & Input to post-hoc                                                 & Score function \\ \hline
                                  Model selection (Sec.~\ref{sec:model_selection_framework}) & \sml{Image features, output of all candidate models $\{f_k\}_{k=1}^K$'s features}        & \sml{Index of the best model (\ref{index pred})}  \\ \hline
Offloading (Section \ref{sec:offloading_framework})                        & \sml{Image features, output of client model $f_0$'s features    }              &\sml{ Performance metric improvement over $f_0$ (\ref{eqn:score_offload})}           \\ \hline
Dataset shift  (Section \ref{sec:domain_adaptation})               & \sml{Image features, output of model after adaptation $f_c$'s features}  & \sml{Performance on the new domain (\ref{eqn:domain_adaptation_score})} \\ \hline
\end{tabular}
\caption{Our general franework encompasses different use cases by simply changing the input features and output score function definitions. 
}
\label{table:summary}
\end{table}

To refine this formulation, we consider the fact that the output $f$ of the neural network might be very high-dimensional, resulting in a less interpretable $g$ as well as being prone to overfitting during the training process. Additionally, the output $f$ may be of variable length (for example, based on the number of objects detected by the black-box model).
Instead of using the full output $f(\x)$, we may only need a subset of the output to use as inputs to $g$. 
Thus, it is reasonable to utilize only a subset of elements from $f(\x)$ as the inputs to $g$. 
On the other hand, some of the original input data $\x$ might also be useful to train $\hat{g}$.
So overall, given a set of features $\text{ftr}(f,\x)$, 
we define the following generalization of \eqref{score pred}:
\begin{align}
\hat{g}=\arg\min_{g\in \Gc}\frac{1}{n}\sum_{i=1}^n\ell(\sm(\y_i,f(\x_i)),g\circ \text{ftr}(\x_i,f)).\label{feature eq}
\end{align}
\noindent This is the form of $\hat{g}$ that we use throughput this paper.
 Later on in Section \ref{sec:numerics}, we explore which features are most useful in learning $\hat{g}$.
 Defining the score functions and performance metrics in this general way allows our framework to cover a variety of use cases.
 We next discuss a simple example of how confidence calibration is a special case of our framework, before showing how our framework applies to three more complex use cases: model selection, device-server offloading, and dataset shift.
 The use cases are summarized in Fig.~\ref{fig:diagram}.

\textbf{Generalized model calibration.} A standard performance metric is the binary classification error $m(\y,\yh)=\onebb_{\y\neq \arg\max_j \yh_j}$, where $\yh_j$ is the $j^\text{th}$ element of $\yh$. In this scenario, $\hat{g}$ aims to predict the accuracy of $f$. \inc{To accomplish this, \eqref{score pred} can be optimized with a Bayes-consistent loss function such as cross-entropy as discussed in \cite{guo2017calibration} to output an estimate of the correct probability $\Pro(\onebb_{\y= \arg\max_j \yh_j}\bgl \x)=\hg\circ f(\x)$.}

More generally, suppose the metric $m$ takes values in $[0,1]$ and we output an estimate $\hat{m}=\hg\circ f$ by solving \eqref{score pred}. We can then introduce calibration errors for $m$ similar to confidence. For instance, the continuous Expected Calibration Error takes the form $\text{ECE}(\hat{m})=\E_{(\x,\y)}[|\E[m(\x,\y)|\hat{m}(\x)=\alpha]-\alpha|]$, which is the average mismatch between the predicted and actual performance. Section \ref{sec:numerics} describes how solving problem \eqref{score pred} leads to refined calibration outputs for metrics specific to object detection. 


\subsection{Use Case: Model Selection}
\label{sec:model_selection_framework}


We next show how our framework can incorporate multiple models for the model selection use case. Besides $f\triangleq f_0$, suppose we have multiple candidate models $F=(f_k)_{k=0}^K$.
Understanding the best model for an input $\x$ can be useful in a variety of scenarios, for example if the inference runs in the cloud with a library of models available~\cite{zhang2017live,zhang2018awstream}, and we need to choose the best model to run.
Specifically, we define the index of the best model as 
\[
\sm_\text{model} (\y,\x,F)=\arg\max_{0\leq k\leq K} m(\y,f_k(\x)).
\]
\inc{In this use case, we set the score function to be $s(\{m_i\}_{i=0}^K)=\arg\max_{0\leq k\leq K} m_i$ with $m_i=m(\y,f_k(\x))$.}
We solve the index prediction problem by setting $\sm\gets \sm_\text{model}$ in (\ref{feature eq})
\begin{align}
\hat{g}=\arg\min_{g\in \Gc}\frac{1}{n}\sum_{i=1}^n\ell(\sm_\text{model}(\y_i,\x_i,F),g\circ \text{ftr}(\x_i,F)).\label{index pred}
\end{align}
The predicted $\hat{s}$ obtained from $\hat{g}$ is the index of the (predicted) best model.


\subsection{Use Case: Device-Server Offloading}
\label{sec:offloading_framework}

Another use case is for a resource-constrained end device (\eg a smartphone) that wants to choose between a smaller local DNN model, or the library of larger models available in the cloud~\cite{chen2019deep}.
Here, let $f_0$  be the lightweight inference-time efficient model that is run on the client device.
$f_1$ is a more accurate model that is more computationally intensive and hence run on in the cloud.
If $f_0$ on the mobile device gives an inaccurate result, ideally we would like to offload the data to the cloud for inference by the stronger models.
Thus the question is:
Can we predict when $f_0$ fails, and if the data should be offloaded; and if so, which server model $f_k$ should be used?
Note that the difference between this offloading use case and the previous model selection use case (Section \ref{sec:model_selection_framework}) is that we only have the result of $f_0$ as input to $\hat{g}$, whereas for model selection $\hat{g}$ has knowledge of all the models $f_0, f_1, \ldots, f_K$.

A naive approach is to send images with low confidence from the mobile device to the server~\cite{wang2020surveiledge}.
A good calibration method can improve this naive approach through accurate confidence estimation; however, confidence alone as the offloading criterion is insufficient, because in some instances offloading even a low-confidence data sample to the server doesn't help (shown later through our experiments).
Therefore, we define the offloading score $\sm_\text{offload}$ as the \emph{difference} between the performance of the mobile device's model, and the best server model
\begin{align}
\sm_\text{offload}(\y,\x,F)=\max_{0\leq k\leq K}m(\y,f_k(\x))-m(\y,f_0(\x))\quad \label{eqn:score_offload}
\end{align}
\noindent We then use $\sm_\text{offload}$ in (\ref{feature eq}) to train $\hat{g}$.


\noindent\textbf{Threshold policy:} After solving for $\hat{g}$ using (\ref{feature eq}) and (\ref{eqn:score_offload}), to utilize the performance estimate, we propose a simple thresholding policy as a proof of concept.
This empirical threshold is used to decide whether to offload new samples. Let $0\leq \rho\leq 1$ be the offloading fraction, \ie the fraction of data items sent for processing on the cloud by one of the models $(f_k)_{k=1}^K$. $\rho=0$ implies that only $f_0$ is used for inference on the mobile device, and $\rho=1$ implies that all data is offloaded to the cloud (which is not a feasible solution due to high latency/communication costs).
Then the threshold is defined as:
\[
s^\rho_\text{threshold}=\lfloor\rho\rfloor\text{'th largest element of }(\sm_\text{offload}(\y_i,\x_i,F))_{i=1}^n.
\]
\noindent 
In other words, we sort the data samples $i$ by their performance gap $s_\text{offload}^m$, and given a desired offloading fraction $\rho$, we can compute the corresponding performance gap $s^\rho_\text{threshold}$.
A new input data item $\x$ is offloaded if its predicted performance gap $\hsm_\text{offload}(\y,\x,F)\geq \max\{s^\rho_{\text{threshold}}, 0\}$.


\subsection{Use Case: Dataset Shift}
\label{sec:domain_adaptation}
In our final example, we consider a dataset shift use case, where the model $f$ has been trained on one domain with data distribution $\Dc_\text{source}\in\Xc\times \Yc$, and is used in a related domain with distribution $\Dc_\text{target}\in\Xc\times \Yc_\text{target}$. 
The motivation is that while existing techniques~\cite{menon2020long} can improve the accuracy on the new domain $\Dc_\text{target}$, they may result in mis-calibrated confidence, hence necessitating post-hoc adjustment. 

We consider two post-hoc models $\ga$ and $\gp$, which are applied sequentially to the outputs of the black box model. 
First, model $\ga$ minimizes the test error in the new domain by using the prediction error as a performance metric, \ie by solving \eqref{score pred} with $\sm(\y,f(\x))=\y$. In many instances, this can be accomplished through logit adjustment to account for changes in class priors \cite{menon2020long}, although more sophisticated strategies might be needed depending on the amount of dataset distribution shift.
$\ga$ is not the focus of this work, as adapting to dataset shift is a well-studied problem~\cite{ovadia2019can}; rather, we focus on performance estimation after dataset shift through model $\gp$.


Following this adaptation to the new domain, we ask the question: Can we accurately predict the inference performance on the new domain using a post-hoc model $\gp$?
To do this, we define the score function for use in (\ref{feature eq}):
\begin{align}
\sm_\text{domain}(\y,f(\x)) = m(\y,\ga\circ f(\x)). \label{eqn:domain_adaptation_score}
\end{align}

\noindent This is perhaps the most straightfoward score function compared to the other use cases, as the score function is simply equal to the performance metric $m$, albeit on the test set in the new domain $\Dc_\text{target}$.


\section{Numerical Experiments}\label{sec:numerics}
\label{sec:experiments}


In this section, we evaluate our post-hoc model \inc{framework} through the three use cases (model selection, mobile offloading, and dataset shift) described in Section \ref{sec:framework}, with different performance metrics. 
We first describe our setup (Section \ref{sec:setup}), our experiments with a single model including the dataset shift use case (Section \ref{sec:local_prediction}), and finally our experiments with multiple models (Section \ref{sec:multiple_models_eval}), including the offloading and model selection use cases.
In summary, our results show that both gradient boosting and neural network-based post-hoc models uniformly outperform the baseline confidence calibration. Among these, gradient boosting trained with handcrafted input features attains the best performance in most scenarios, as opposed to full-dimension input features, suggesting that more information is not always better.
\subsection{Experiment Setup}
\label{sec:setup}


\begin{table}
\small
	\centering
	\begin{tabular}{l|p{0.85cm}|c}
		\hline
		\textbf{Object detection model} & \textbf{Dataset} & \textbf{ECE}\\
		\hline
		 SSD MobileNets~\cite{liu2016ssd} & COCO&0.01648\\
		 SSD ResNet-50~\cite{liu2016ssd} & COCO&0.01657\\
		 SSD Inception~\cite{liu2016ssd} & COCO&0.01748\\
		 Fast R-CNN ResNet-101~\cite{girshick2015fast} & COCO&0.01976\\
		 Tiny YOLO~\cite{huang2018yolo} & VOC&0.01847\\
		 YOLO v2~\cite{redmon2017yolo9000} & VOC& 0.01790\\
		\hline
	\end{tabular}
	\caption{Models used in our evaluation. Local prediction: The ECE of Post-hoc-XGB for F1-score prediction is low across models and datasets, demonstrating the generality of our approach across different black-box models. }

	\label{tab:model_list}
\end{table}

\textbf{Datasets and pre-trained models:}
We evaluate our framework using two types of machine learning tasks: image classification and object detection. 

\emph{Image classification:} We use CIFAR-10~\cite{krizhevsky2009learning}, which consists of 60,000 32x32 color images in 10 classes, with 6000 images per class.
We split the dataset into training, two validation, and test sets, with 4500/100/300/1000 images per class, respectively.
The target dataset $\Dc_\text{target}$ for the dataset shift use case (V1, V2) is a modified version of the two validation sets.
Essentially, to model the distribution shift on the new domain, we randomly select 3 out of the original 10 classes and sample these classes with frequencies 3:3:1), then V1 is used to train the dataset shift calibrator $f_c$, and V2 is used to train the post-hoc model. 
The black-box machine learning model is a ResNet model~\cite{he2016deep}. 

\emph{Object detection:} We use the VOC and COCO datasets. VOC~\cite{Everingham15} contains 11,540 images with objects from 20 target classes.
We randomly pick 1000/500 images for validation and test, respectively.
COCO~\cite{lin2014microsoft} contains 91 classes, with 2.5 million labeled instances in 328k images in the dataset. We use 4000/1000 images for validation and test, respectively.
We evaluate 9 different pre-trained models on these datasets, ranging from compressed models designed for mobile devices to more powerful models, as summarized in Table \ref{tab:model_list}.

\begin{table}[]
\small
\centering
	\begin{tabular}{|l|p{6cm}|}
	\hline
	\multirow{3}{*}{\makecell{\textbf{Image}\\\textbf{features}}}        &  {\textbf{Color histogram entropy}}                        \\ \cline{2-2} 
														& {\textbf{Image size}}                           \\ \cline{2-2} 
														& {\textbf{Number of corners}}            \\ \hline
	\multirow{8}{*}{\makecell{\textbf{Model}\\\textbf{features}}} 		&  \textbf{Class score}, representing the importance of a class towards $m$; computed as the weights from a linear regression between the \# of objects per class and $m$.              \\ \cline{2-2} 
											                              & \textbf{Location score}, representing the importance of a location in an image towards $m$; computed as the weight from a linear regression between 25 grid locations and $m$.
											                              \\ \cline{2-2} 
														 & \textbf{Min confidence} across objects in an image          \\ \cline{2-2}
														                                & \textbf{Max confidence} across objects in an image          \\ \cline{2-2} 
														                                & \textbf{Mean confidence} across objects          \\ \cline{2-2} 
														                                & \textbf{\# of bounding boxes} in an image \\ \cline{2-2} 
														                                & \textbf{Min bounding box size} across objects   \\ \cline{2-2} 
														                                & \textbf{Mean bounding box size} across objects  \\ \hline
	\end{tabular}
	\caption{Handcrafted features used in our post-hoc model.
	}
		\label{tab:features}
\end{table}	 

\textbf{Post-hoc model:} For the post-hoc model, we experimented with two model designs: a 3-layer fully connected neural network and gradient boosting using XGBoost~\cite{chen2016xgboost}.
We label the former as \emph{\textbf{Post-hoc-NN}} throughout our experiments, and the latter as \emph{\textbf{Post-hoc-XGB}}.
The neural network has 2 hidden layers. 
The number of neurons is two thirds of the input size for each layer (\eg for dataset shift, the input size is 20, so the number of neurons per layer is 13 and 9 respectively).
The activation function is a ReLu layer. The learning rate is set to 0.03.
The gradient boosting method uses trees with maximum depth 5, subsampling ratio 0.7, learning rate 0.1 and numer of epoch 300.
We varied these hyperparameters and chose the values above based on their overall performance.

\textbf{Per-image features:} 
 The inputs to the post-hoc model $\gh$ include model features (outputs of the black-box machine learning model) and image features (features pertaining to the original input image), as shown in Table~\ref{tab:features}.
We created handcrafted features to summarize these inputs. 
The model features require summarization because the black-box model outputs might be very high-dimensional and their dimension is not fixed.
The image features similarly require summarization because the high image resolution results in high-dimensional features.
The intuition behind per-image features, considering the multiple models use case and number of bounding boxes feature as an example, is that knowing that there are many bounding boxes suggests a more complex image, suggesting that the post-hoc model should predict that a more powerful machine learning model is needed.
In Section \ref{sec:object_detection}, we systematically investigate which of the per-image features correlate best with the performance metrics.

We note that, different from previous work that uses per-object features
\cite{Kuppers_2020_CVPR_Workshops,schwaiger2021black}, we use per-image features (\ie an aggregation of the features of all the objects in an image). This allow us to overcome the influence of undetected ground truth.
In other words, using per-object features alone results in undetected objects, whose information can't be incorporated into the performance estimate (since they are false negatives and never detected).
 Per-image features, as we use, contain general information of the whole image, so it contains information about those undetected objects and can help us estimate per-image performance metrics such as recall and F1-score.

\textbf{Metrics:} 
The overall evaluation metrics include: 
\begin{itemize}
	\item \textbf{Expected calibration error (ECE)}~\cite{naeini2015obtaining,guo2017calibration}:
	The ECE measures the calibration accuracy, by sorting the predictions into $J$ bins ($J=10$ in our experiments), and counting how many samples were put into the correct bins, weighted by the empirical probability of that bin:
	$\mathrm{ECE}=\sum_{j=1}^{J} \frac{\left|B_{j}\right|}{n}\left|\operatorname{true}\left(B_{j}\right)-\operatorname{predict}\left(B_{j}\right)\right|$,
	where $n$ is the number of samples, $\operatorname{true}(B_{j})$ is the number of samples that fall in bin $B_j$, and $\operatorname{predict}(B_{j})$ is the number of samples that were predicted to fall into bin $B_j$. ECE are used in generally throughout Section \ref{sec:local_prediction}, as it is one of the most popular metrics to evaluate calibration accuracy.
	\item \textbf{Coefficient of determination ($R^2$):} The $R^2$ value is a measure of correlation, defined as $1-\frac{\sum_{i}\left(y_{i}-\hat{y}\right)^{2}}{\sum_{i}\left(y_{i}-\hat{y}_{i}\right)^{2}}$, where $y_i$ is true value, $\bar{y}$ is the mean value of y, and $\hat{y}_i$ is the predicted value.  This metric is used in Figure~\ref{fig:correlation} to show the correlation between features and performance metrics.

	\item \textbf{Spearman's rank correlation coefficient:} The Spearman correlation coefficient is defined as the Pearson correlation coefficient between two rankings.
	In our case, we compute the Spearman correlation coefficient between the ground truth ranking and the ranking produced by the post-hoc model (e.g., convert the predicted F1 scores to a ranking). This is used in Section \ref{sec:object_detection} and Section \ref{sec:domain_adaptation_experiments} for the basic evaluation and the dataset shift use case.
\end{itemize}

Different evaluation metrics are appropriate for different scenarios.
Although the ECE is widely used to evaluate calibration, it has several drawbacks, leading us to consider the other additional metrics listed above.
First, when the prediction output is continuous, the ECE computation requires bins, which are artifically introduced just for evaluation purposes.
Second, ECE is sensitive to arbitrary bin settings, (\eg the more number of bins we use, the larger the error).
Finally, we sometimes care more about whether one sample performs better than another, rather than the exact value of the performance metric. In the offloading use case, for example, we want to know which sample has better performance on the server compared to its peers, rather than exactly how much better it will perform. 
Rank correlation gives us the accuracy of these rankings, without being influenced by arbitrary settings, such as the number of bins for the ECE calculation.

The metrics we use to evaluate object detection and image classification include:
\begin{itemize}
	\item \textbf{Intersection over Union (IoU)}~\cite{Rezatofighi_2018_CVPR}: IoU measures how well a detected object's location matches with the ground truth (for those objects correctly classified), and is defined as:
	$\text{IoU} =\frac{|A \cap B|}{|A \cup B|}$
	where A and B are areas of the predicted and ground truth bounding boxes, respectively.
	\item \textbf{F1-score, recall, and precision:} F1-score is a measure of accuracy of the classification task, defined in terms of recall ($\frac{\text{true positives}}{\text{true positives + false negatives}}$) and precision ($\frac{\text{true positives}}{\text{true positives + false positives}}$) as $2 \cdot \frac{\text { precision } \cdot \text { recall }}{\text { precision }+\text { recall }}$.
	We compute the F1-score per image.
	\item \textbf{Accuracy:} The classification accuracy is defined as $\frac{1}{n} \sum_i 1_{\hat{y}_i = y_i}$.
\end{itemize}


\textbf{Baselines:} The baseline methods that we compare our post-hoc model against include:
\begin{itemize}
	\item \textbf{Confidence:}
	For image classification, we use the regular confidence values output by the machine learning model.
	For object detection, the model may output a both a location confidence and class confidence/probability for each object in an image. 
	For such models (\eg YOLO), we compute the combined confidence by multiplying the class probability (first term) and the location confidence (second and third terms) as:
	$P\left( \text{class} \mid \text{object} \right) * P(\text{object}) * IOU$.
	For models without location confidence, such as SSD, we just use the class confidence.
	\item \textbf{Calibrated confidence:} Confidence calibration methods include temperature scaling and vector scaling \cite{guo2017calibration}.
	Briefly, temperature scaling scales all the logits by a scalar parameter $T$, while vector scaling is a multi-class extension that applies a linear transformation to the logits.
To extend this to object detection, which also cares about an object's location, when training the calibration model we label a detected object as having 0 confidence if there is no object actually present.
\end{itemize}


\subsection{Performance Prediction of a Single Model}
\label{sec:local_prediction}

\subsubsection{Base Case: Local Prediction}
\label{sec:object_detection}
     




We first study in-depth whether our post-hoc model works well to predict performance of object detection locally on a device.
Success in this base scenario is necessary before applying our framework to more complex use cases in the following subsections.
We examine the impact of different setups, including feature selection, post-hoc model performance, generality to different black-box models and datasets, and sample complexity.

 \begin{figure}
	\centering
	\begin{subfigure}[t]{0.4\textwidth}
				\includegraphics[width=\linewidth]{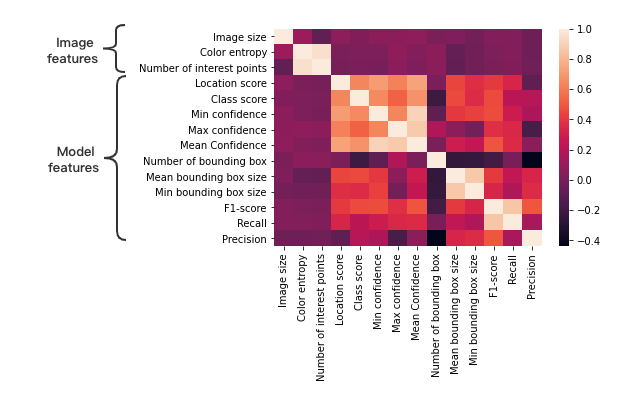}\vspace{-6pt}	
			\end{subfigure}
			\vspace{-0.2cm}
	\caption{Correlation between the features and the performance metrics (F1-score, recall, precision). The black-box model features are generally more highly correlated with the performance metrics than image features. 
	}
	\label{fig:correlation}
\end{figure} 

\textbf{Feature selection:} First, we investigate which hand-crafted features to use as inputs to the post-hoc model.
We find that the features relating to the black-box model, rather than image features, are more highly correlated with the performance estimates, and thus more useful to the post-hoc model.
This is shown in Figure~\ref{fig:correlation}, which illustrates the correlation between the handcrafted features and the performance metrics (F1-score, recall, precision). The feature selection depends on the size of dataset, cases and object detection/image classification models. For instance,  the class score and location score are highly correlated with the performance in normal scenarios with large enough validation set to train the post-hoc model, but can hurt if there is not enough training data (discussed later in the sample complexity experiments).


\begin{figure}
	\centering
	\begin{subfigure}[t]{0.15\textwidth}
		\includegraphics[width=\linewidth]{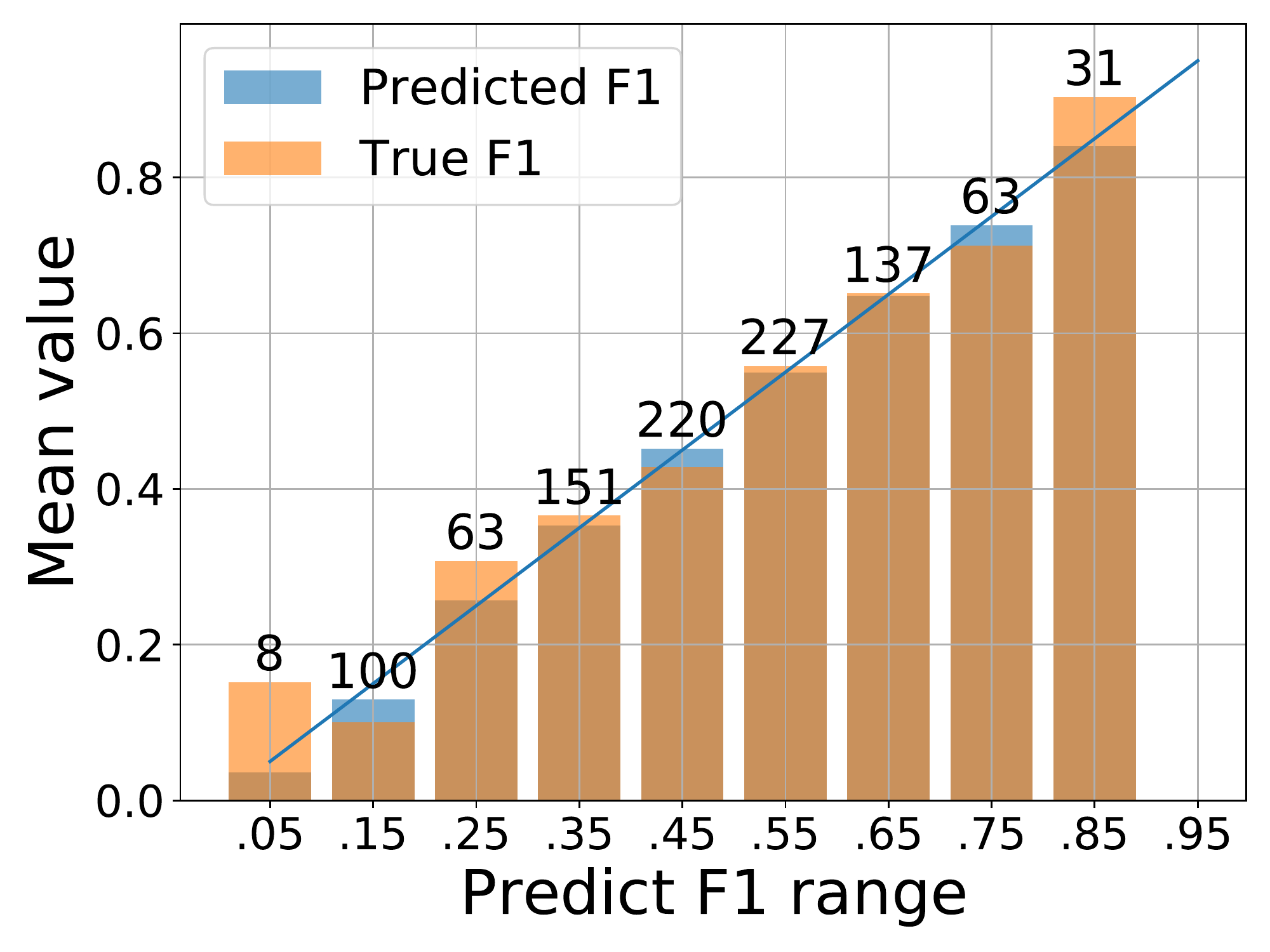}\vspace{-6pt}
		\caption{Post-hoc-NN (logit features) 
		}\label{fig1a}		
	\end{subfigure}
	~
	\begin{subfigure}[t]{0.15\textwidth}
		\includegraphics[width=\linewidth]{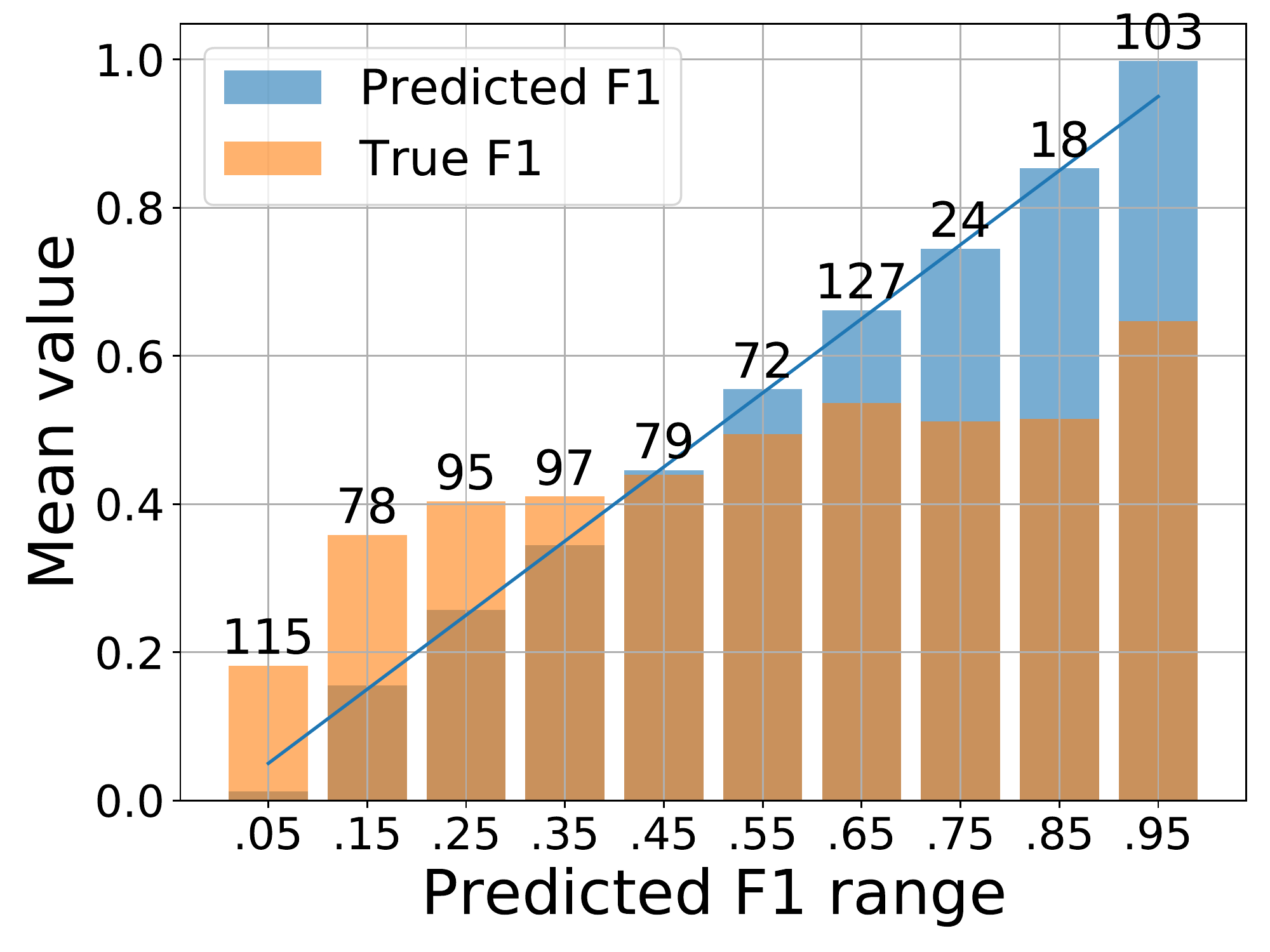}\vspace{-6pt}
		\caption{Post-hoc-NN (handcrafted features)}\label{fig1b}
	\end{subfigure}
	~
	\begin{subfigure}[t]{0.15\textwidth}
		\includegraphics[width=\linewidth]{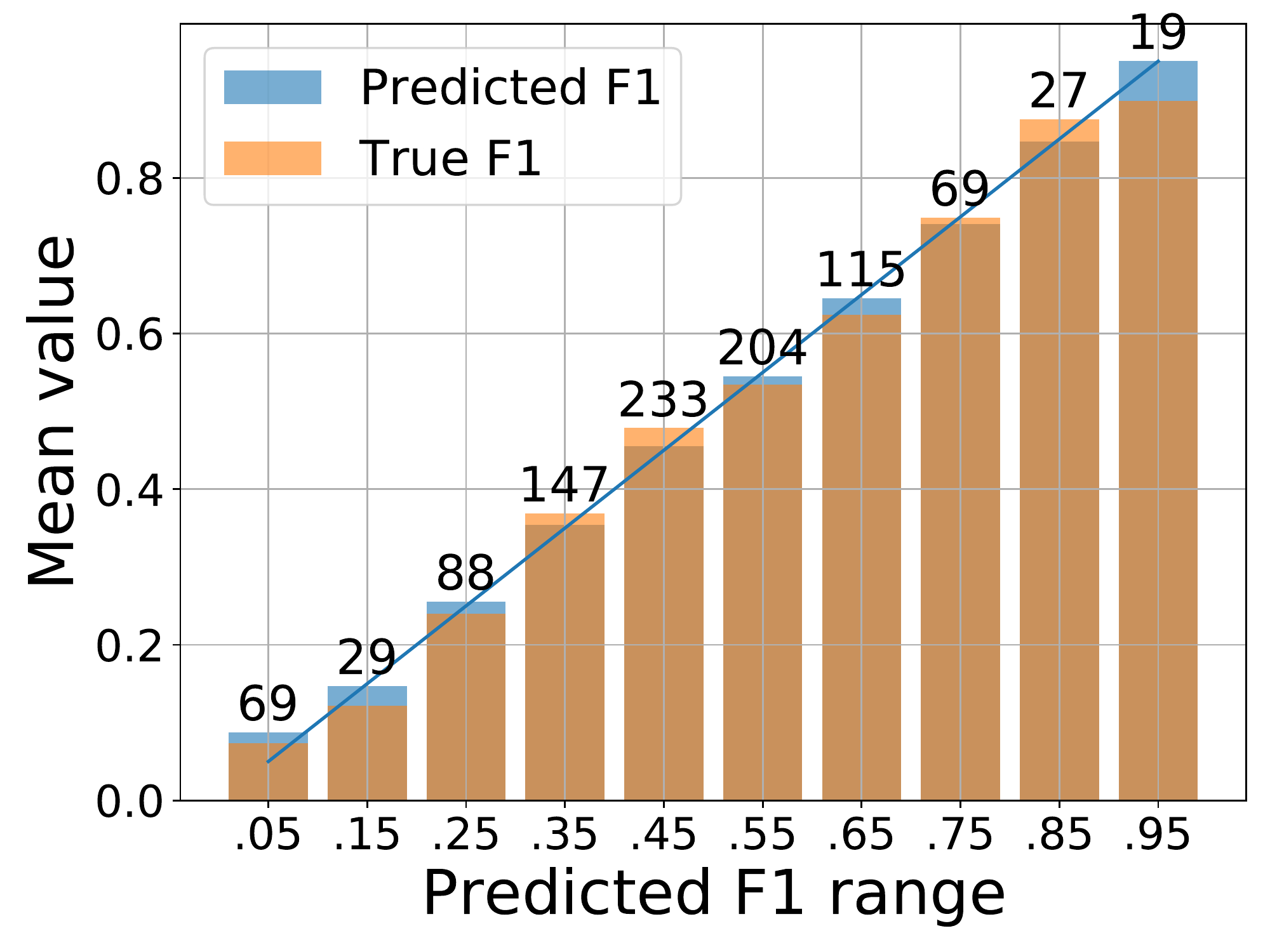}\vspace{-6pt}
		\caption{Post-hoc-XGB (handcrafted features)}\label{fig1c}
	\end{subfigure}\\
	~
	\begin{subfigure}[t]{0.15\textwidth}
		\includegraphics[width=\linewidth]{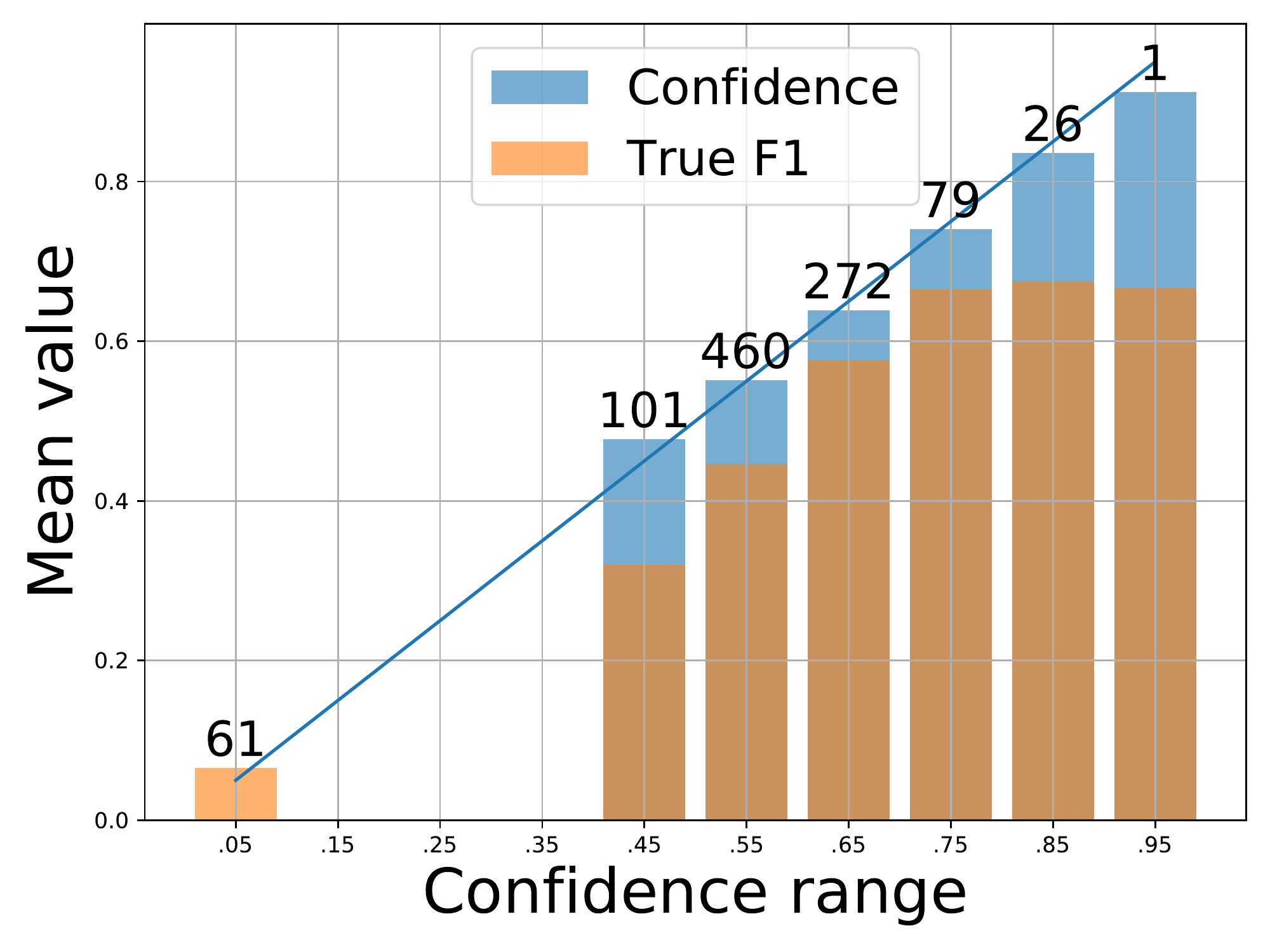}\vspace{-6pt}
		\caption{Confidence}\label{fig1d}
	\end{subfigure}
	~
	\begin{subfigure}[t]{0.2\textwidth}
		\includegraphics[width=0.75\linewidth]{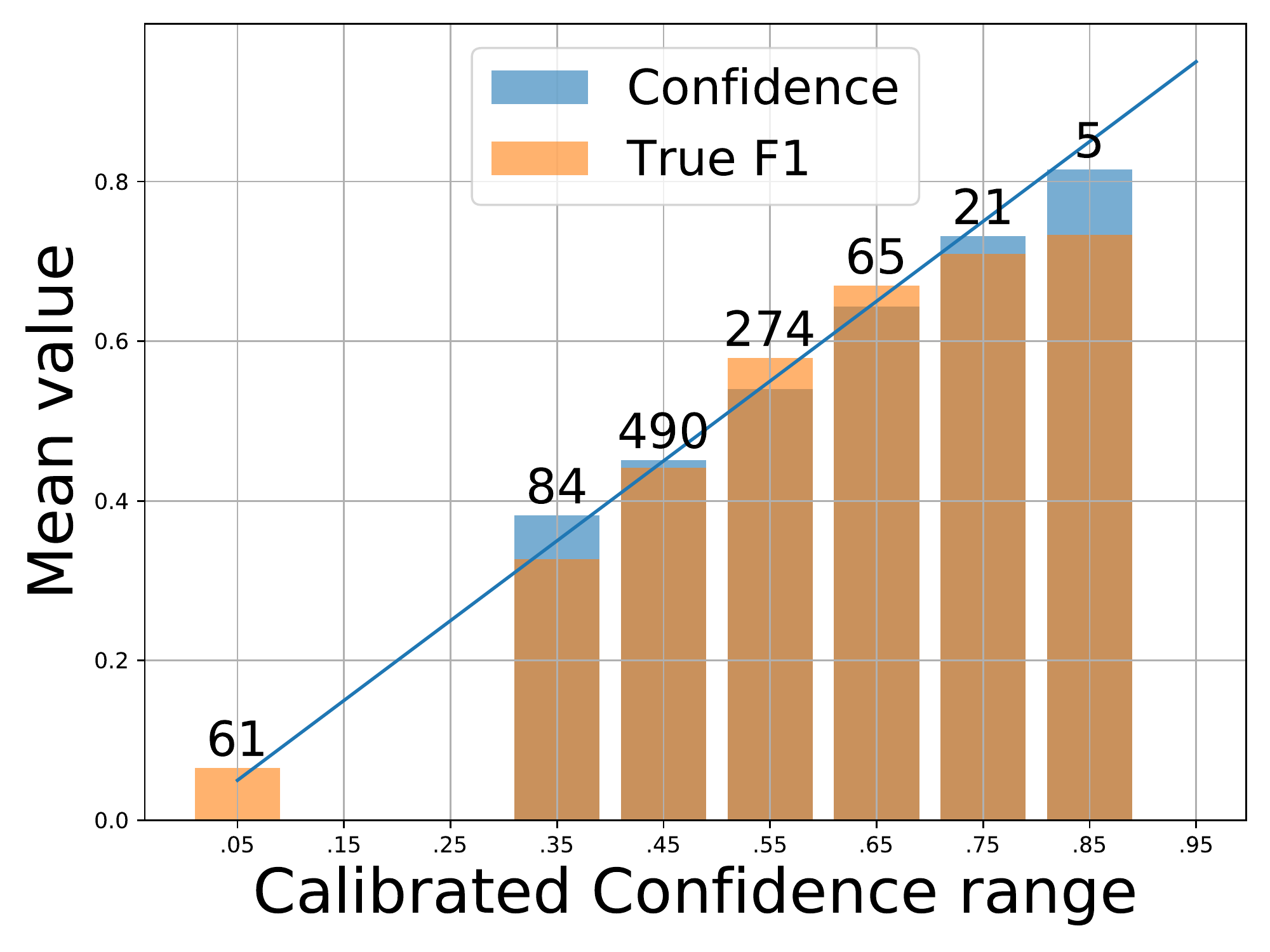}\vspace{-6pt}
		\caption{Calibrated confidence}\label{fig1e}
	\end{subfigure}
	\caption{Local prediction: Reliability diagrams, generated by binning the test examples by F1-score predictions from our post-hoc models (a,b,c), confidence (d) and calibrated confidence (e), and plotting the average predicted value (blue) and average true value (orange).
	With perfect prediction, the true value should align exactly with the predicted value along the diagonal. 
	}
	\label{fig:diff_models}
\end{figure} 

\textbf{Post-hoc model performance:} Using these handcrafted features, we next evaluate the performance of the post-hoc model.
We trained a post-hoc model to predict three different performance metrics -- precision, recall, and F1-score -- of the object detector SSD MobileNets. To evaluate the performance of post-hoc model, we compute ECE to show the the difference in expectation between the predicted value and the true object detection model's performance metrics. Taking F1-score as an example, the calibrated confidence has an ECE of around \clr{0.08}, while our post-hoc model (specifically, Post-hoc-XGB) can decrease ECE to less than 0.02 (latter shown in Table~\ref{tab:model_list}).

To further examine this performance gain, in Figure~\ref{fig:diff_models}, we plot the reliability diagrams of our approach and the baselines.
These reliability diagrams show the mean value of the metric in question, as a function of the predicted metric.
Figures~\ref{fig:diff_models}(a), (b), and (c) show the different post-hoc model variants as a predictor of F1, while (d) and (e) show confidence as a predictor of F1.
XGBoost with handcrafted features as input (Figure~\ref{fig:diff_models}(a)) performs the best, as it most closely tracks the diagonal. 

\begin{figure}
	\centering
	\begin{subfigure}[t]{0.15\textwidth}
			\includegraphics[width=\linewidth]{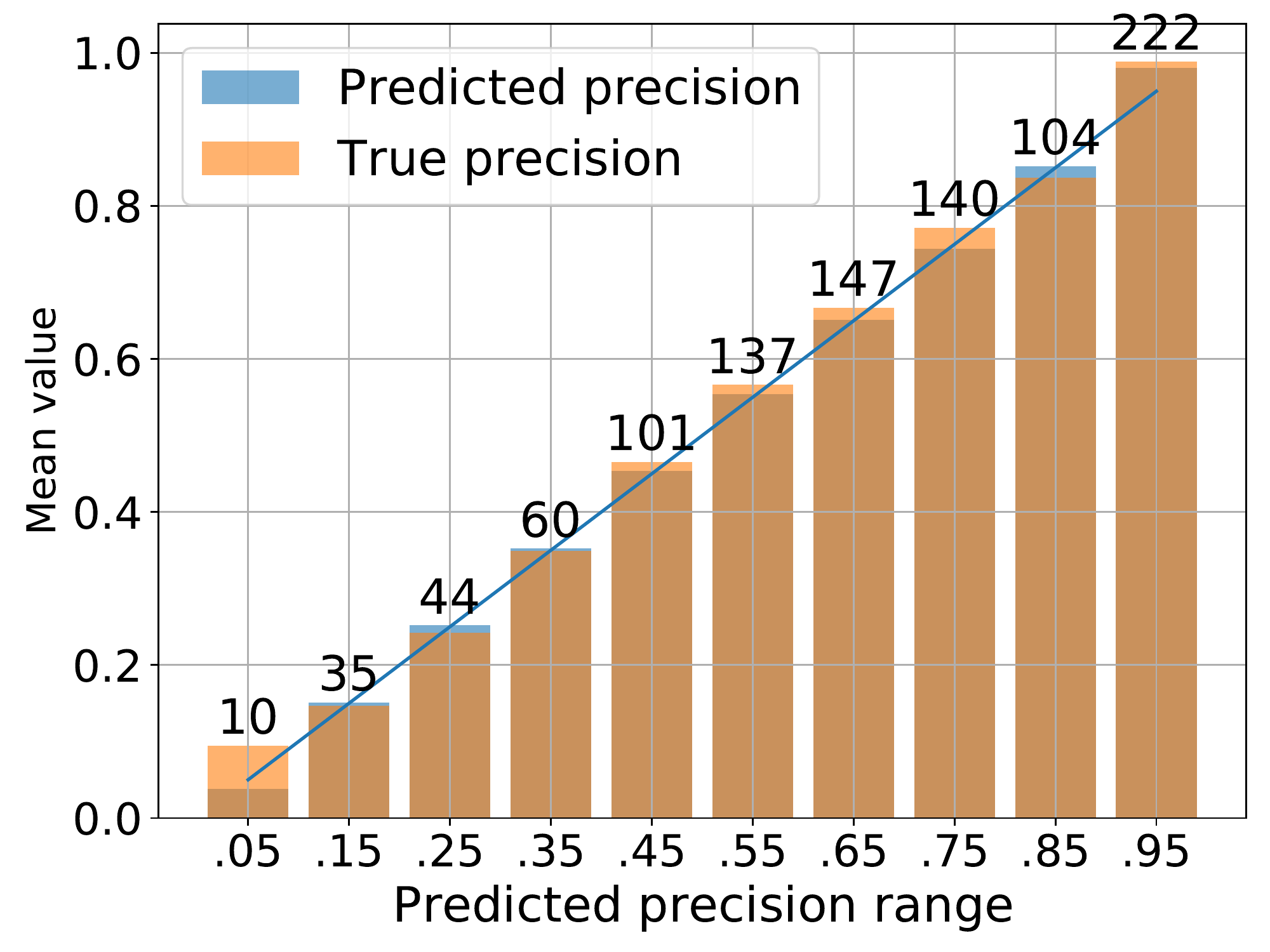}\vspace{-6pt}
			\caption{Precision}\label{fig1a}		
		\end{subfigure}
		~
		\begin{subfigure}[t]{0.15\textwidth}
			\includegraphics[width=\linewidth]{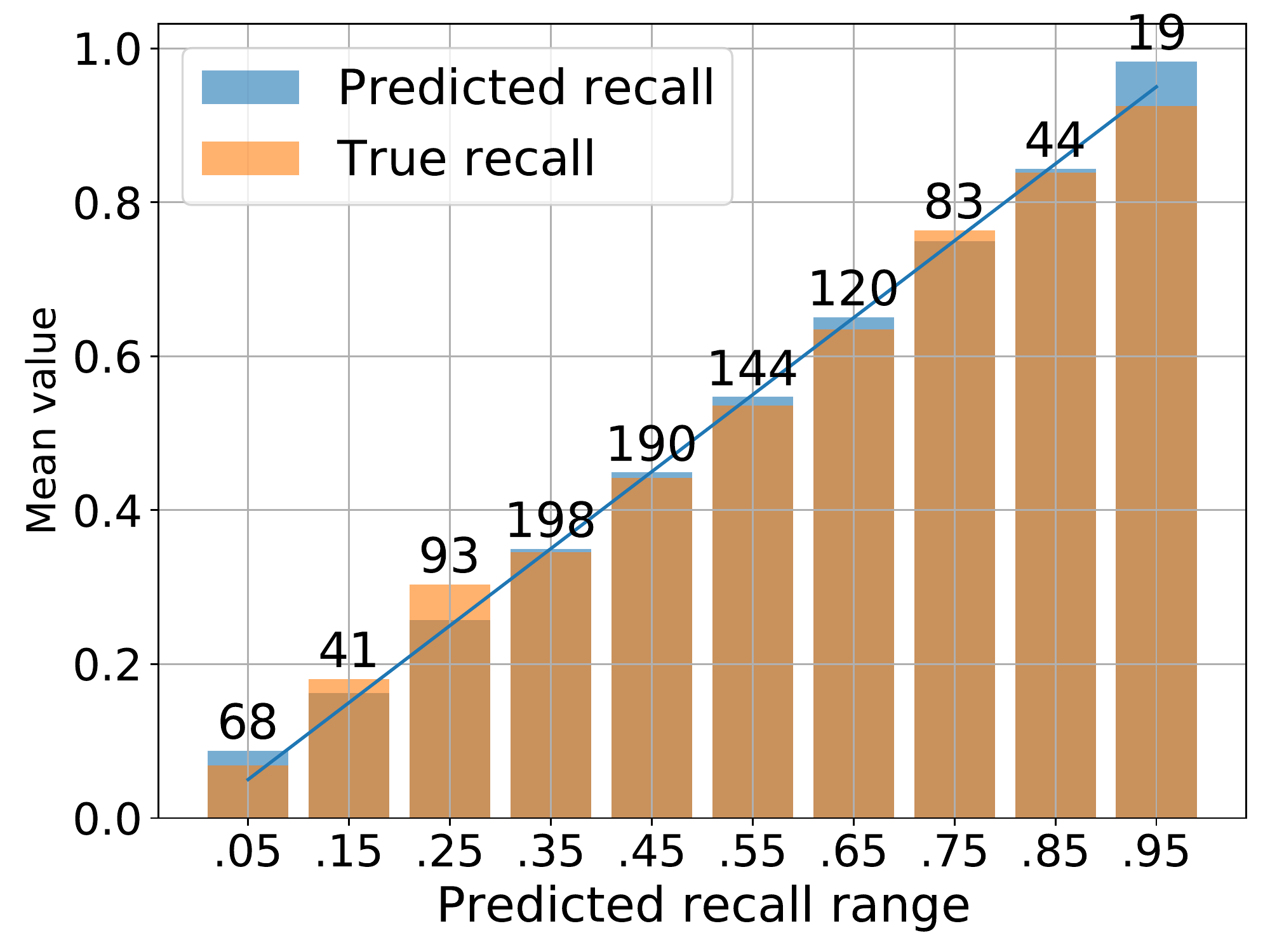}\vspace{-6pt}
			\caption{Recall}\label{fig1b}
		\end{subfigure}
		~
		\begin{subfigure}[t]{0.15\textwidth}
			\includegraphics[width=\linewidth]{Figures/1_feature_f1_xgboost_bucket}\vspace{-6pt}
			\caption{F1-score}\label{fig1c}
		\end{subfigure}
		
	\caption{Local prediction: Reliability diagram of Post-hoc-XGB for different performance metrics, generated by binning the test examples by predicted (a) precision, (b) recall, and (c) F-1 score.}
	\label{fig:nn_metrics}
\end{figure}

Figure~\ref{fig:nn_metrics} shows the ability of Post-hoc-XGB to predict other performance metrics (precision and recall); the results indicate that our model can successfully predict these other performance metrics. 

\textbf{Generality across models and datasets:}
 The results of our experiments with five additional object detection models and one additional dataset are shown in Table~\ref{tab:model_list}, leading to similar conclusions on the good performance of Post-hoc-XGB. 
Although different models have different structures and principles, and are trained for different datasets, our post-hoc model can estimate their performance accurately.

\emph{In summary, our results show that XGBoost with handcrafted features can accurately predict F1-score, precision, and recall, and outperforms the baselines across 6 different black-box models and 2 datasets. 
}

\begin{figure}
\centering
	\begin{subfigure}[t]{0.23\textwidth}
			\includegraphics[width=\linewidth]{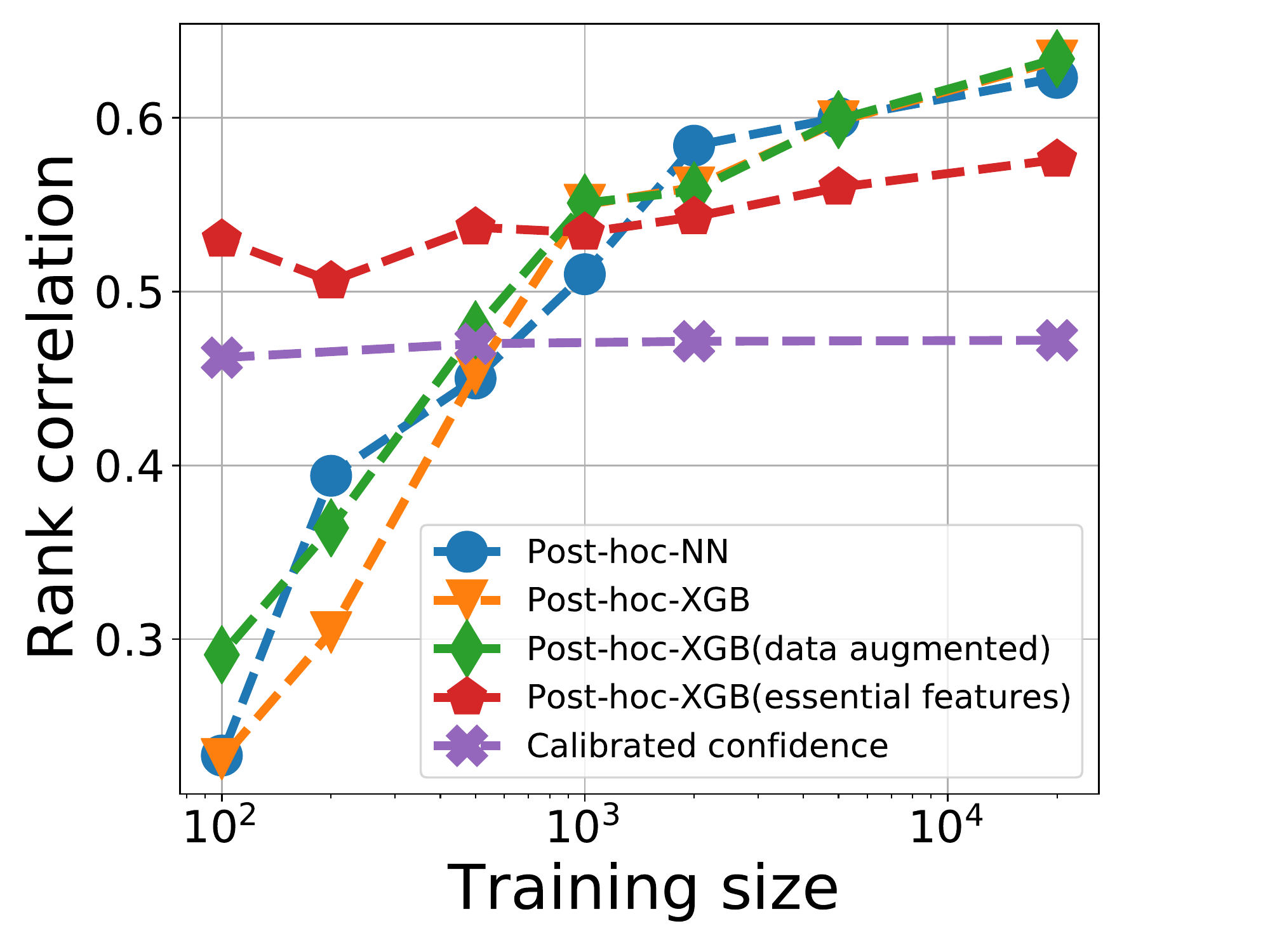}\vspace{-6pt}
			\caption{Rank correlation}\label{fig1a}		
		\end{subfigure}
		~
		\begin{subfigure}[t]{0.23\textwidth}
			\includegraphics[width=\linewidth]{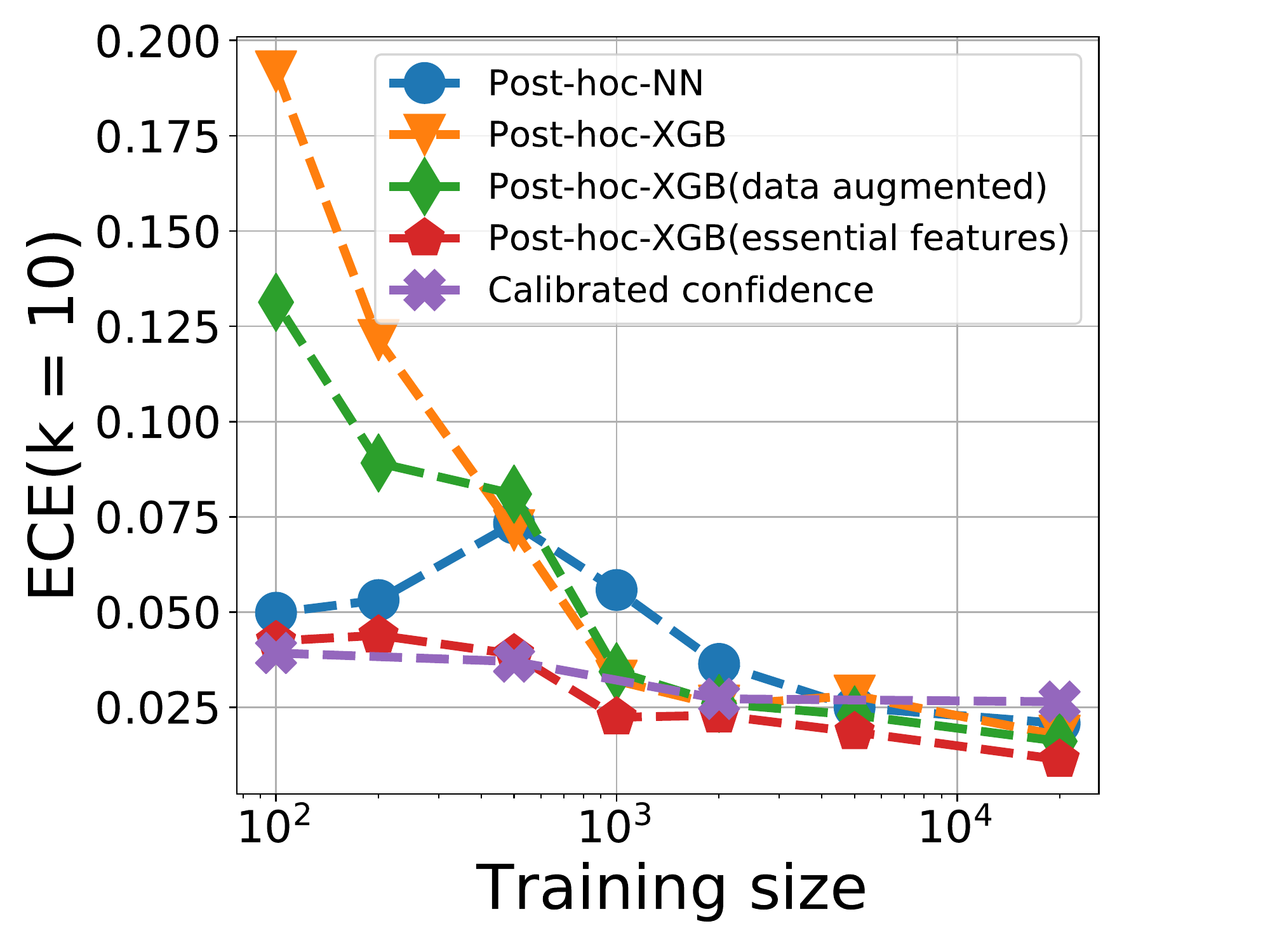}\vspace{-6pt}
			\caption{ECE}\label{fig1b}
		\end{subfigure}
		
	\caption{Local prediction: We investigate the sample complexity for predicting F1-score. Post-hoc training is done for varying sizes of validation dataset. As sample size increases, more complex model tends to perform better both for ECE and rank correlation. 
	}
	\label{fig:sample_complexity}
\end{figure}

\textbf{Sample complexity:}
We also experiment with how much data is needed to train the post-hoc model and the impact of a small training dataset.
In Figure~\ref{fig:sample_complexity}, we show the ECE and the rank correlation for the post-hoc model trained with different dataset sizes (\ie the validation set size).
To counteract the effect of small training set sizes, we introduce two new variants of the post-hoc model: ``Post-hoc-XGB (essential features)'' and ``Post-hoc-XGB(data augmented)''.
The former excludes class and location score from the handcrafted features, as those scores are 
harder to infer from when there are few samples; 
the latter adds data augmentation (e.g., cropping, rotating) when training the post-hoc model.

The results show that while data augmentation is beneficial for small sample size its impact is rather limited (comparing the green curves to the orange curves). This makes sense as the post-hoc model needs to analyze what is contained each image (\eg number of objects, object classes) in order to make good predictions, and simply rotating or cropping images doesn't give the post-hoc model more training examples with more diverse contents (\eg more objects or diverse object classes).
Overall, as long as we can select the handcrafted features correctly (i.e., use Post-hoc-XGB (essential features) when the training set is small, and the regular Post-hoc-XGB when the training set is larger), Post-hoc-XGB can achieve good performance, better than or on par with other baselines across all training set sizes.
The fact that ``Post-hoc-XGB (essential features)'' performs quite well even when there is limited training data demonstrates the algorithmic benefits of feature selection and reducing the
dimensionality of the  post-hoc model's input. 

\subsubsection{Use Case: Dataset Shift}
\label{sec:domain_adaptation_experiments}
\textbf{Setup:} 
We emulate dataset shift by generating new datasets with different class distributions. 
 First, for image classification, we use the CIFAR-10 dataset, which has 10 classes, and a ResNet model. To model the dataset distribution shift on the new domain, 
 we randomly select 3 out of the original 10 classes and sample these classes with frequencies 3:3:1.
Let (V1,V2) be the datasets of the new domain. 
Second, for object detection, a ResNet model is trained on the COCO dataset and the VOC dataset is used as the new domain. V1 and V2 are from the VOC dataset.
Recall that we must train two models, $\ga$ for dataset shift and $\gp$ for performance estimation. To obtain $\ga$, we use vector scaling trained on V1. $\gp$ is trained on V2 using either the handcrafted features for object detection, or for image classification, a slightly modified set of features (logits, the label predicted by the black-box model, and compressed image using PCA)

\begin{figure}
	\centering
	\begin{subfigure}[t]{0.15\textwidth}
		\includegraphics[width=\linewidth]{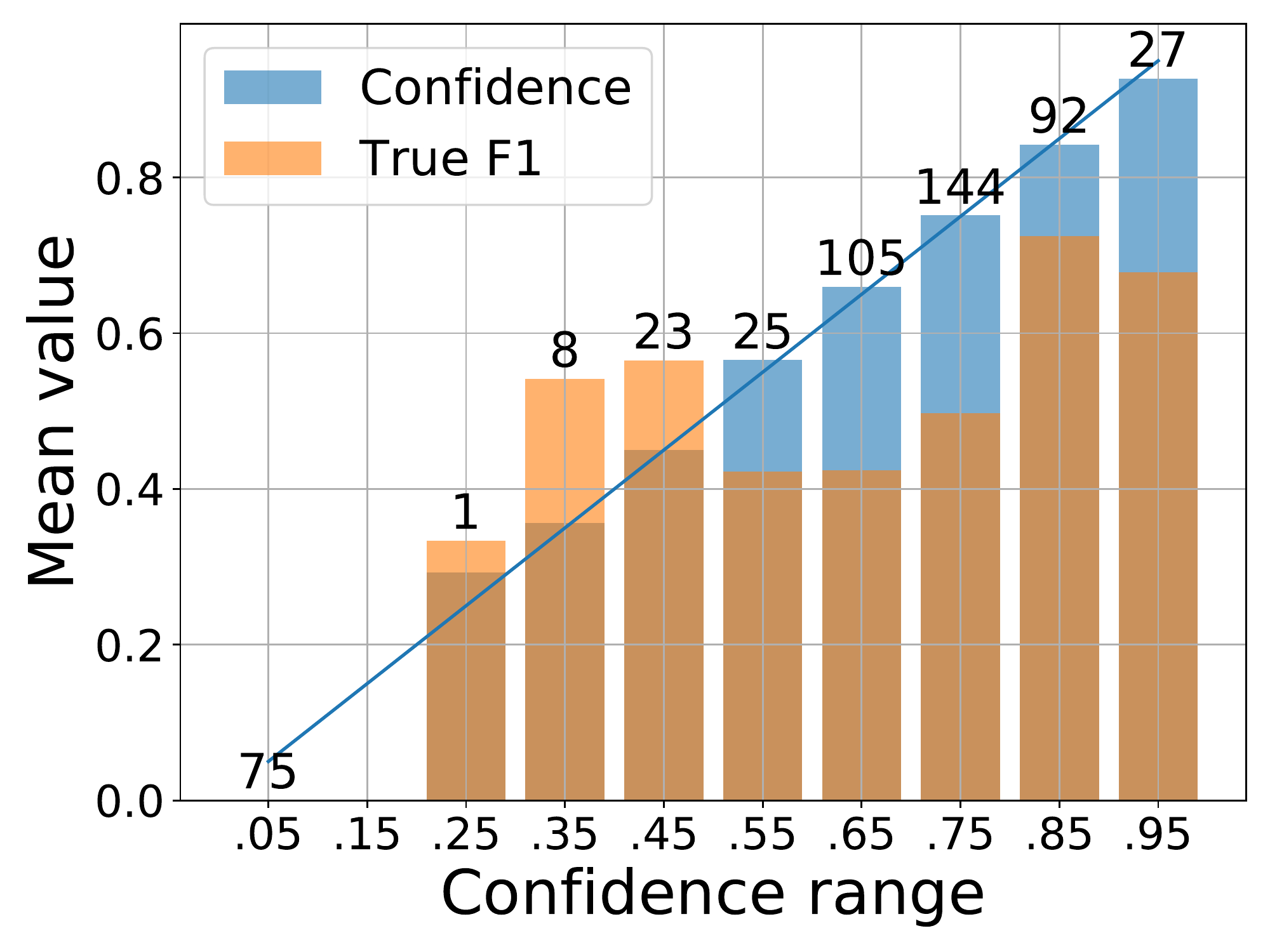}\vspace{-6pt}\label{fig7a}
		\caption{Confidence of $f_c$}		
	\end{subfigure}
	~
	\begin{subfigure}[t]{0.15\textwidth}
		\includegraphics[width=\linewidth]{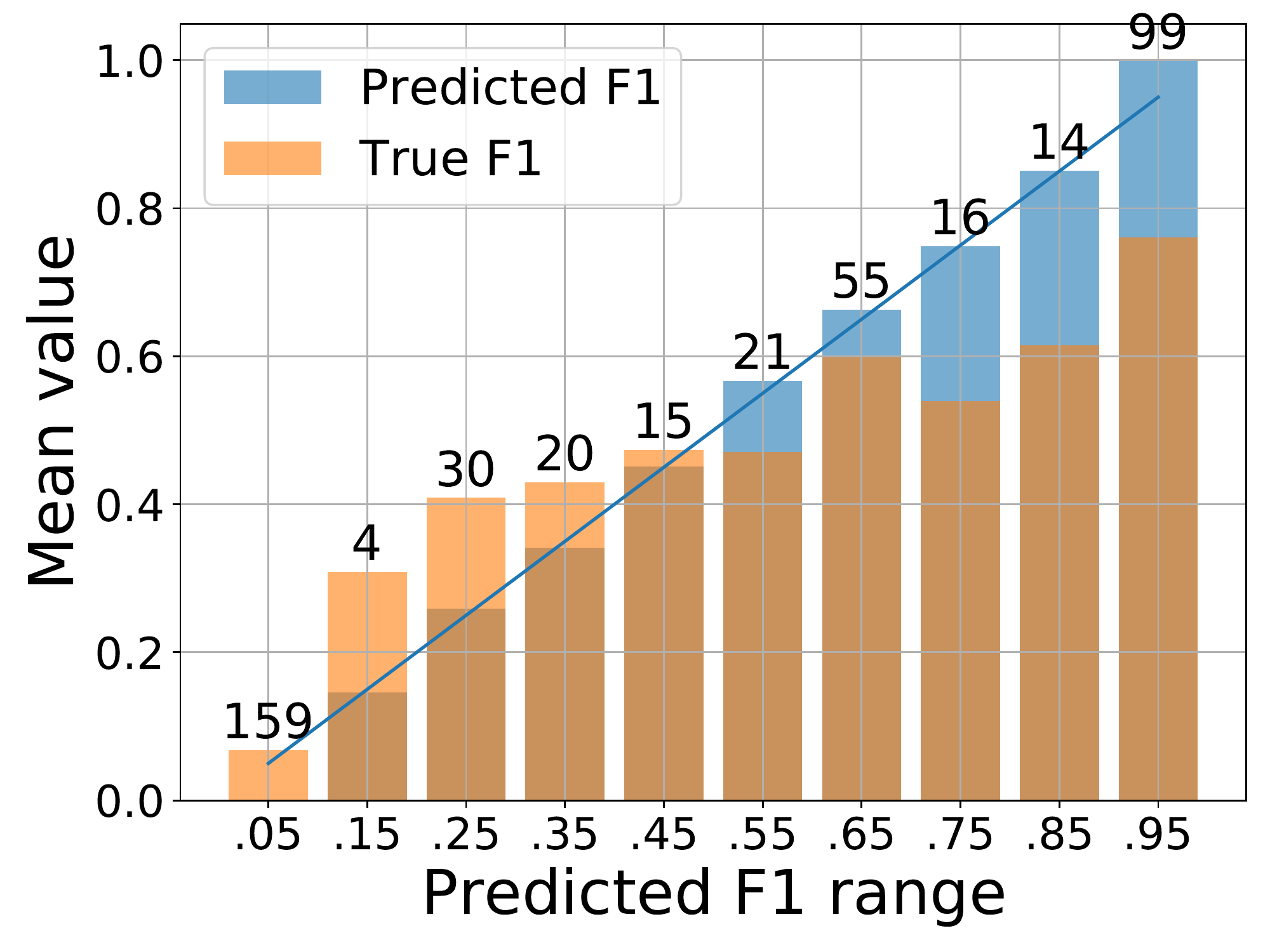}\vspace{-6pt}
		\caption{Post-hoc-NN}\label{fig7b}
	\end{subfigure}
	~
		\begin{subfigure}[t]{0.15\textwidth}
			\includegraphics[width=\linewidth]{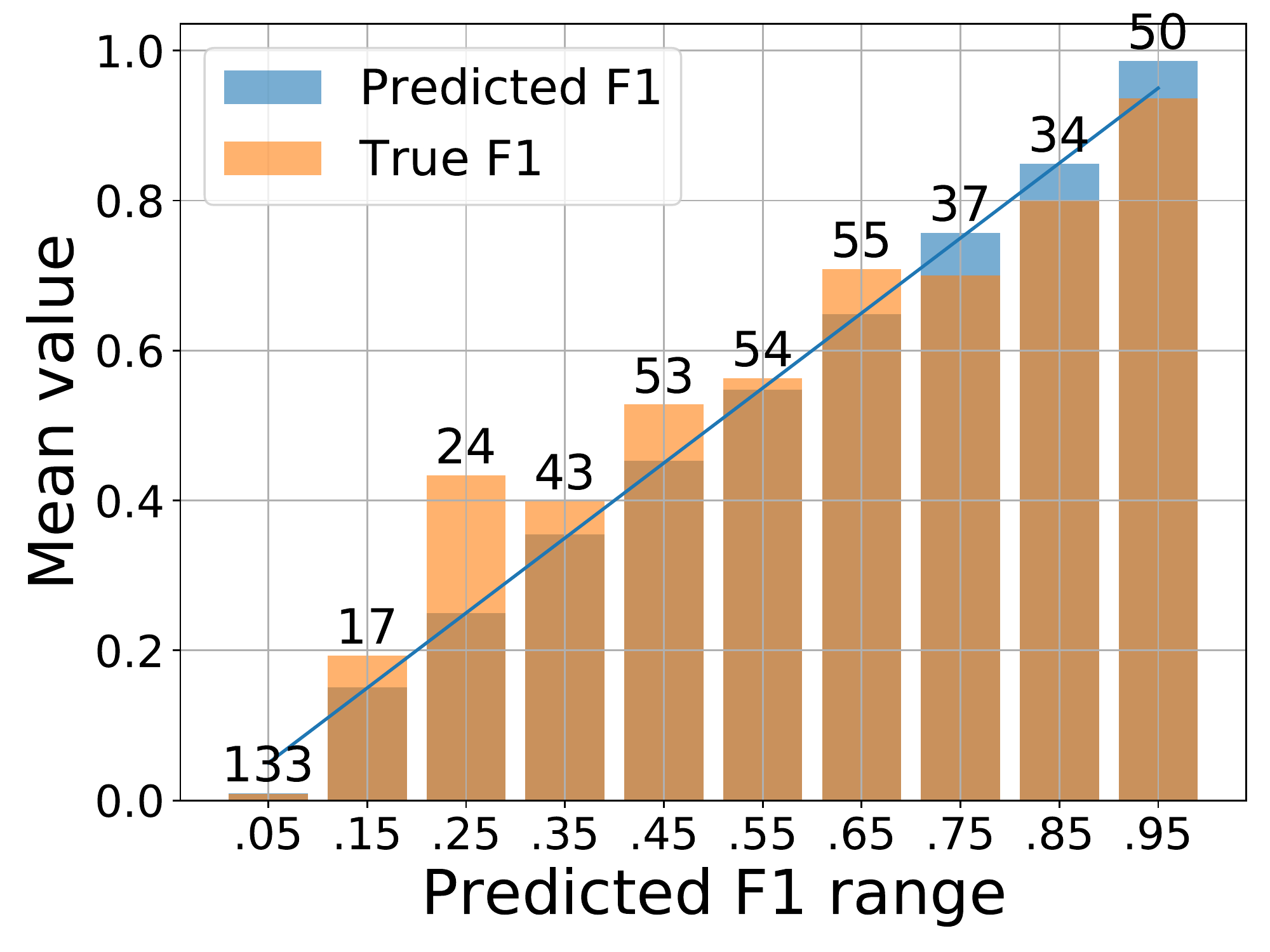}\vspace{-6pt}
			\caption{Post-hoc-XGB}\label{fig7c}
		\end{subfigure}
		\caption{Dataset shift: 
		Reliability diagrams for (a) the confidence of the dataset shift model $f_c$, and (b)(c) the F1-score of our post-hoc models.
		}
		\label{fig:domain_object}
\end{figure}

\textbf{Object detection results:}
Here we focus on the accuracy of the predicted class for each detected object in the image. 
We analyze the performance through reliability diagrams as before.
Without a post-hoc model, the confidence of the original model plus dataset shift $f_c$ is mis-calibrated (Figure~\ref{fig:domain_object}a).
Figures~\ref{fig:domain_object}(b) and (c) show the reliability of our post-hoc model. 
Post-hoc-XGB (Figure~\ref{fig:domain_object}c) achieves the best reliability (closest to diagonal) and thus can be helpful to predict the performance after dataset shift.
\begin{table}
\small
	\centering
	\begin{tabular}{c|c|c}
		\hline
		& \textbf{ECE} & \textbf{Rank correlation} \\
		\hline
		\textbf{Confidence} & 0.17330 & 0.281\\
		\hline
		\textbf{Post-hoc-NN} & 0.11344&0.560\\
		\hline
		\textbf{Post-hoc-XGB} &0.04299 & 0.584\\
		\hline
	\end{tabular}
	\caption{Dataset shift:  Comparison of ECE, MAE, and $R^{2}$ for the confidence baseline and post-hoc models.}
	\label{tab:domain_object_ECE}
\end{table}

Table~\ref{tab:domain_object_ECE} highlights the performance improvement of our post-hoc model compared to the baseline domain calibration method for different metrics, with Post-hoc-XGB having the best (lowest) ECE, and slightly better rank correlation than Post-hoc-NN.
These results are similar to that of the object detection base case discussed previously (Section \ref{sec:object_detection}), and provides further evidence that the simpler XGBoost-based post-hoc model outperforms a more complex post-hoc model based on neural networks.


\begin{figure}
	\centering
	
	\begin{subfigure}[t]{0.15\textwidth}
		\includegraphics[width=\linewidth]{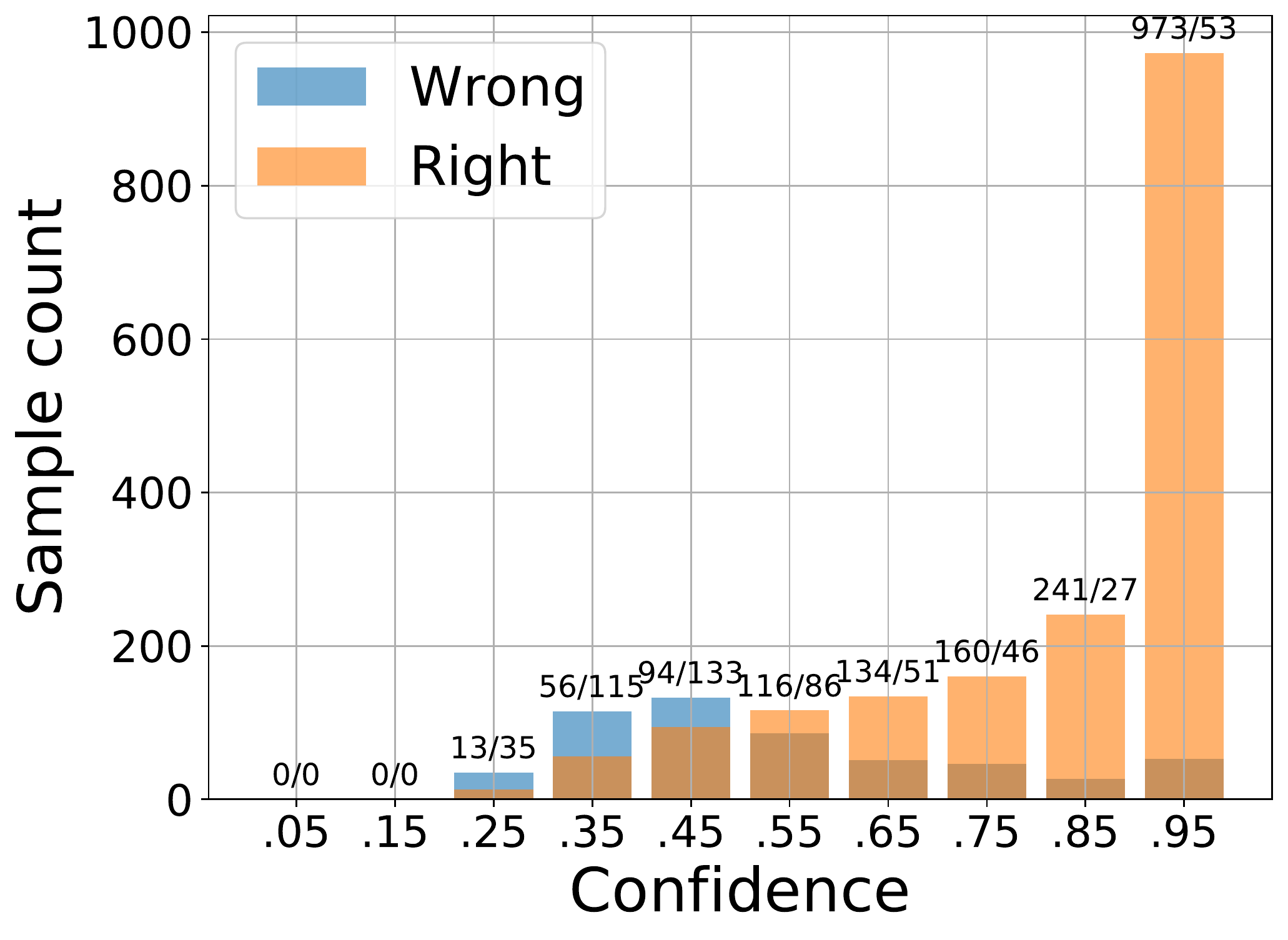}\vspace{-6pt}\label{fig7a}
		\caption{Confidence of $f_c$}		
	\end{subfigure}
	~
	\begin{subfigure}[t]{0.15\textwidth}
		\includegraphics[width=\linewidth]{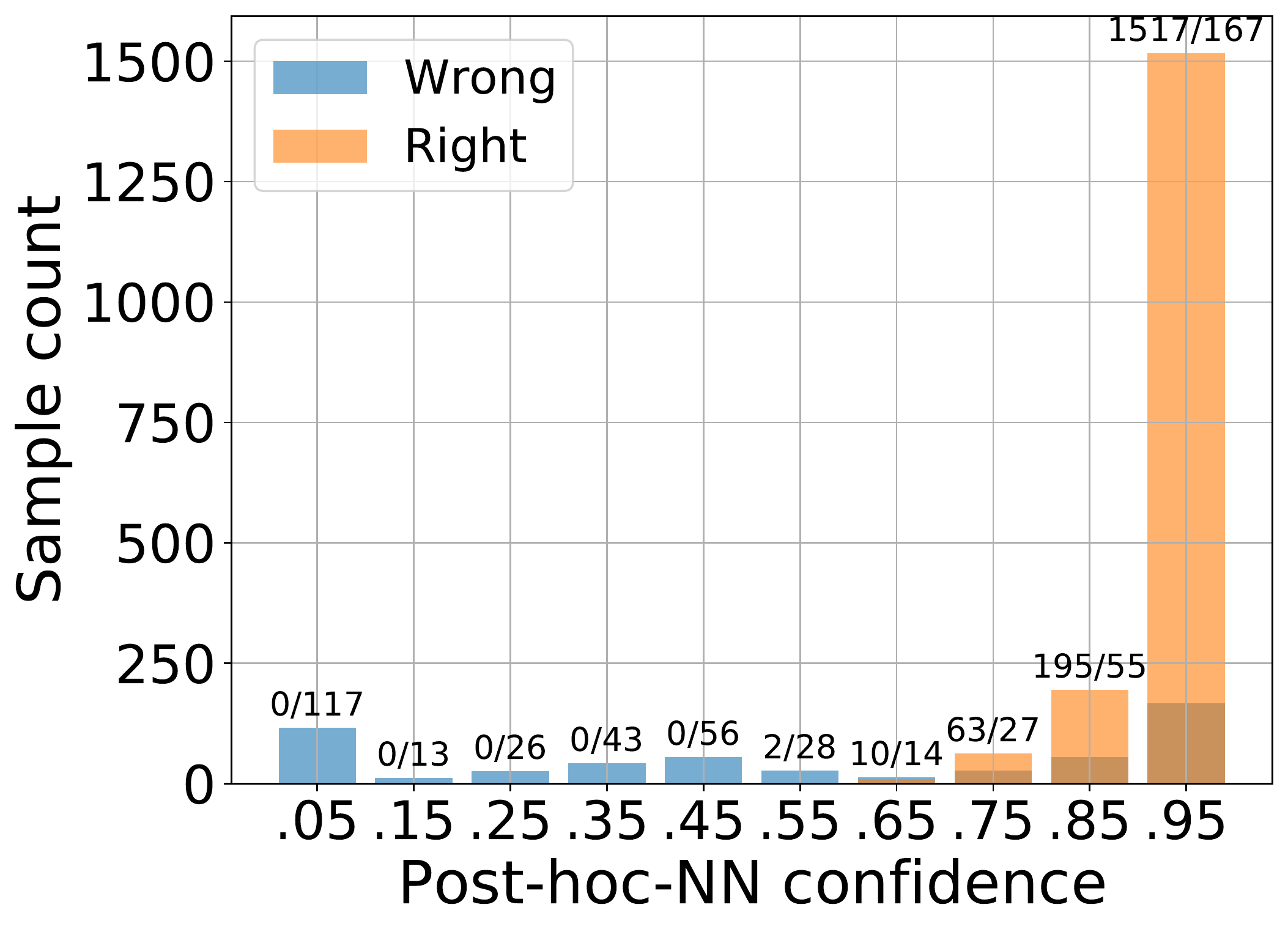}\vspace{-6pt}
		\caption{Post-hoc-NN}\label{fig7b}
	\end{subfigure}
	~
		\begin{subfigure}[t]{0.15\textwidth}
			\includegraphics[width=\linewidth]{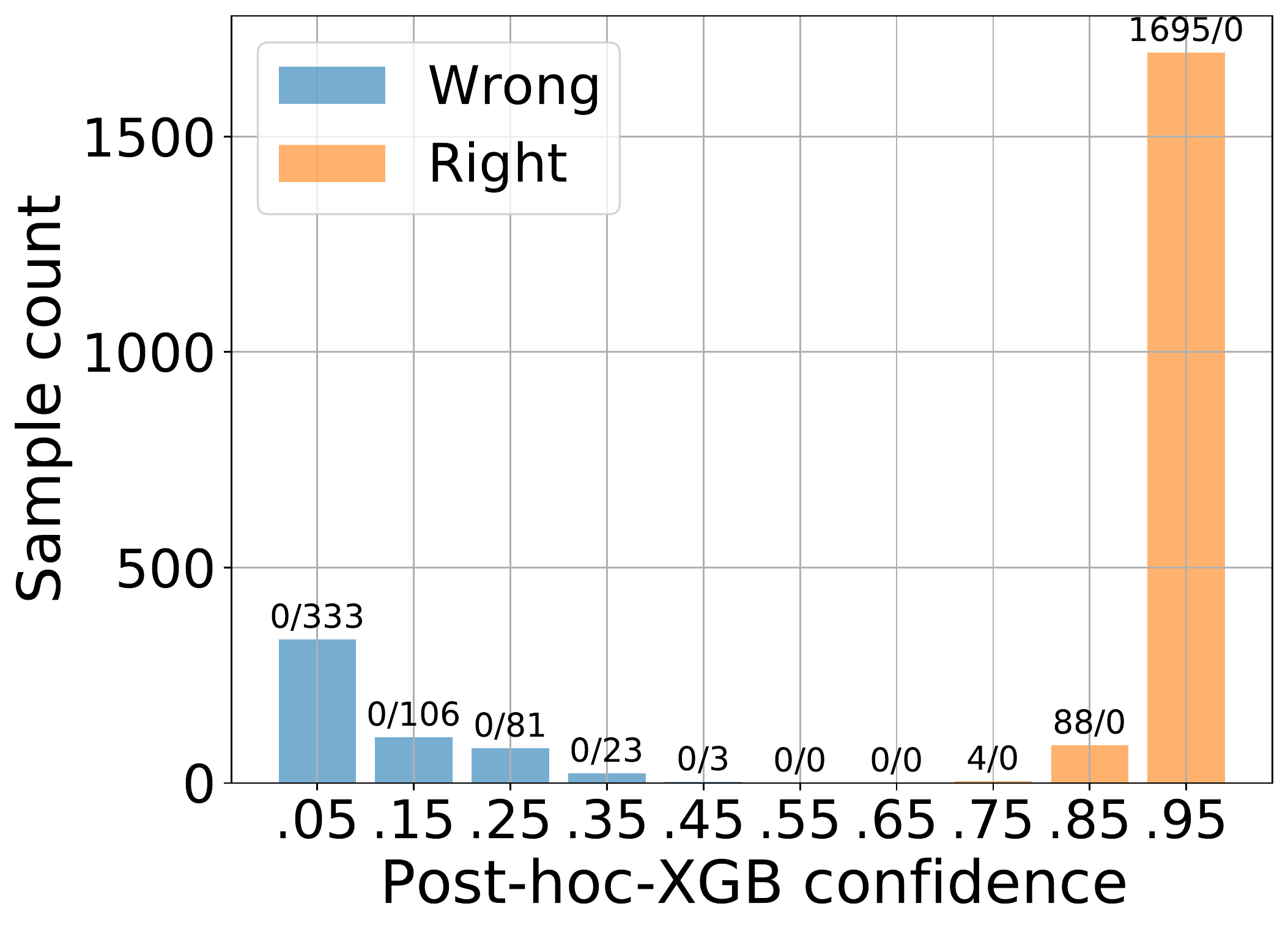}\vspace{-6pt}
			\caption{Post-hoc-XGB}\label{fig7c}
		\end{subfigure}

		\caption{Dataset shift: Distribution of rightly/wrongly classified samples, generated by binning the test samples by confidence, as achieved by the (a) dataset shift model $f_c$ or our post-hoc prediction models (b)(c). Ideally, we should see high confidence samples classified rightly, and vice versa.}
		\label{fig:domain_image_count}
\end{figure}

\textbf{Image classification results:}
The reliability diagram results for image classification are consistent with that of object detection, showing that the post-hoc model can accurately predict the performance metrics after dataset shift, and are therefore omitted for brevity.
Instead, we examine another facet of the problem: whether the post-hoc model can use an estimated performance metric (i.e., confidence) to 
 separate mis-classified samples from the correct ones. 
Ideally, we would expect that a low predicted confidence would correspond to an incorrectly classified sample by the black-box model, and a high predicted confidence corresponds to correctly classified samples.
In other words, is there a threshold for which samples with predicted confidence below the threshold are classified wrongly, while samples above the threshold are classified correctly?

Figure~\ref{fig:domain_image_count} shows the number of rightly/wrongly classified samples for different confidence bins.
Ideally, we would expect all the wrongly classified samples (blue) to be to the left of the figure (low predicted confidence), while all of the correctly classified samples (orange) to be on the right of the figure (high predicted confidence).
We see that post-hoc-XGB predicts very well (Figure~\ref{fig:domain_image_count}c), achieving good separation between the wrongly/correctly classified samples, with the wrongly classified samples corresponding to predicted confidence of less than 0.7, and the correctly classified samples corresponding to predicted confidence larger than 0.5.
In contrast, the vanilla dataset shift model $f_c$ (Figure~\ref{fig:domain_image_count}a) has high-confidence samples that are mis-classified, and vice versa.

\emph{In summary, our post-hoc model works well to predict performance after dataset shift, for both image classification and object detection, decreasing ECE by 0.13 for example, compared to the confidence baseline in object detection. 
} 
\subsection{Performance Prediction of Multiple Models}
\label{sec:multiple_models_eval}
In this section, our goal is to evaluate how well post-hoc models can estimate the performance of a combination of black-box machine learning models.

\subsubsection{Use Case: Device-Server Offloading}

\label{sec:offloading}

In this set of experients, we study whether the post-hoc model can correctly determine the F1-score gap (\ref{eqn:score_offload}) between the mobile device's model and the server's model.
This is challenging because we do not know server model's output in advance to make the prediction, but intuitively, for images that are more complex or with many small objects, the gap will be larger and the image should be offloaded to a server with a more powerful black-box model.
We set $K=1$, \ie there is one machine learning model stored on a client, and one machine learning model available on a cloud server.
 The client and server machine learning models we used are listed in Table~\ref{tab:model_list}; we tried many combinations with similar results, so for brevity, only the results from \clr{SSD MobileNet} paired with \clr{SSD ResNet-50}, as well as SSD Inception paired with Fast R-CNN ResNet-101 are discussed. 

\begin{figure}
	\centering
	\begin{subfigure}[t]{0.2\textwidth}
		\includegraphics[width=\linewidth]{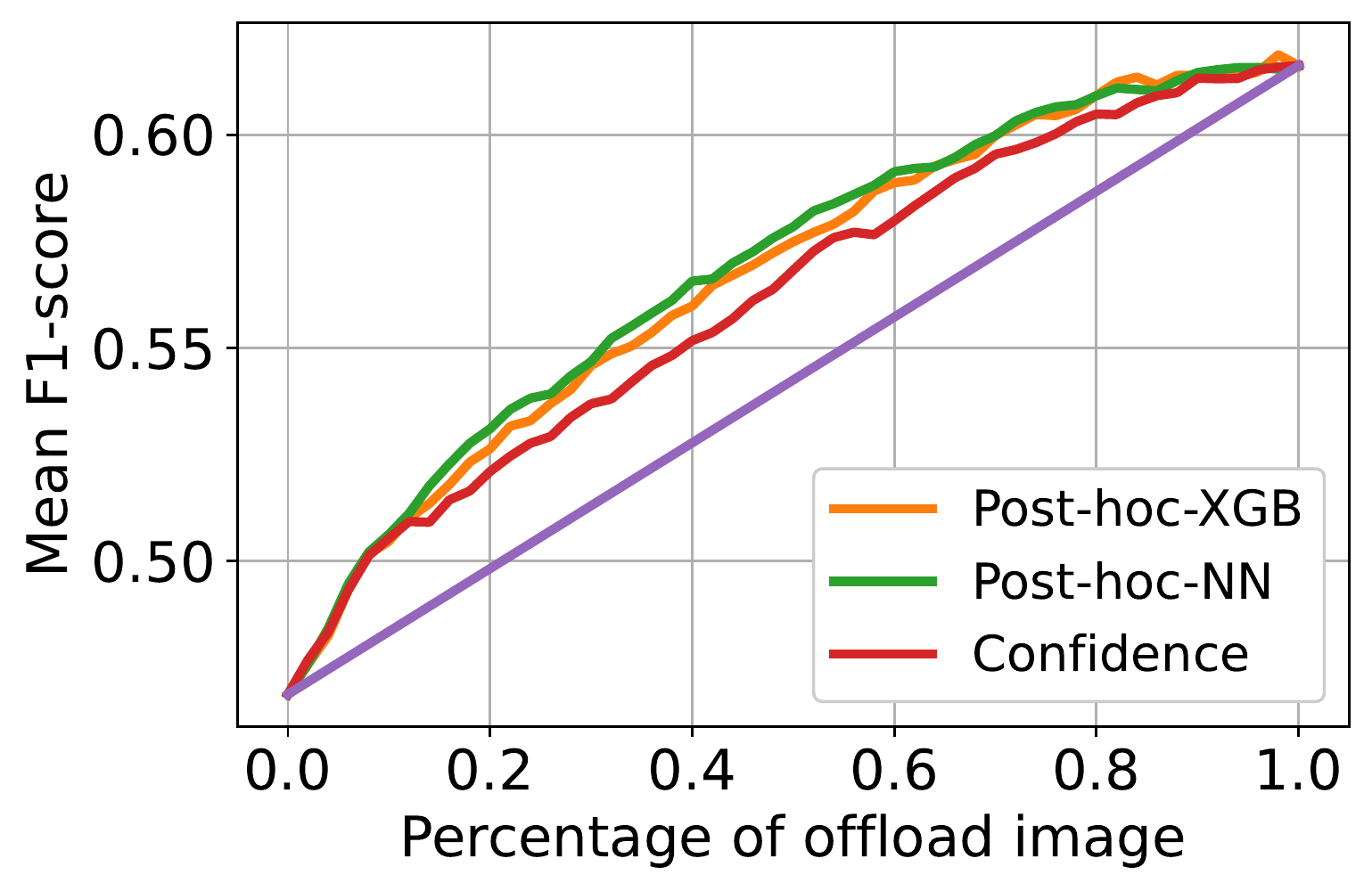}\vspace{-6pt}
		\caption{SSD MobileNet (client) \& SSD ResNet-50 (server) }\label{fig3a}
	\end{subfigure}	
	~
		\centering
		\begin{subfigure}[t]{0.2\textwidth}
			\includegraphics[width=\linewidth]{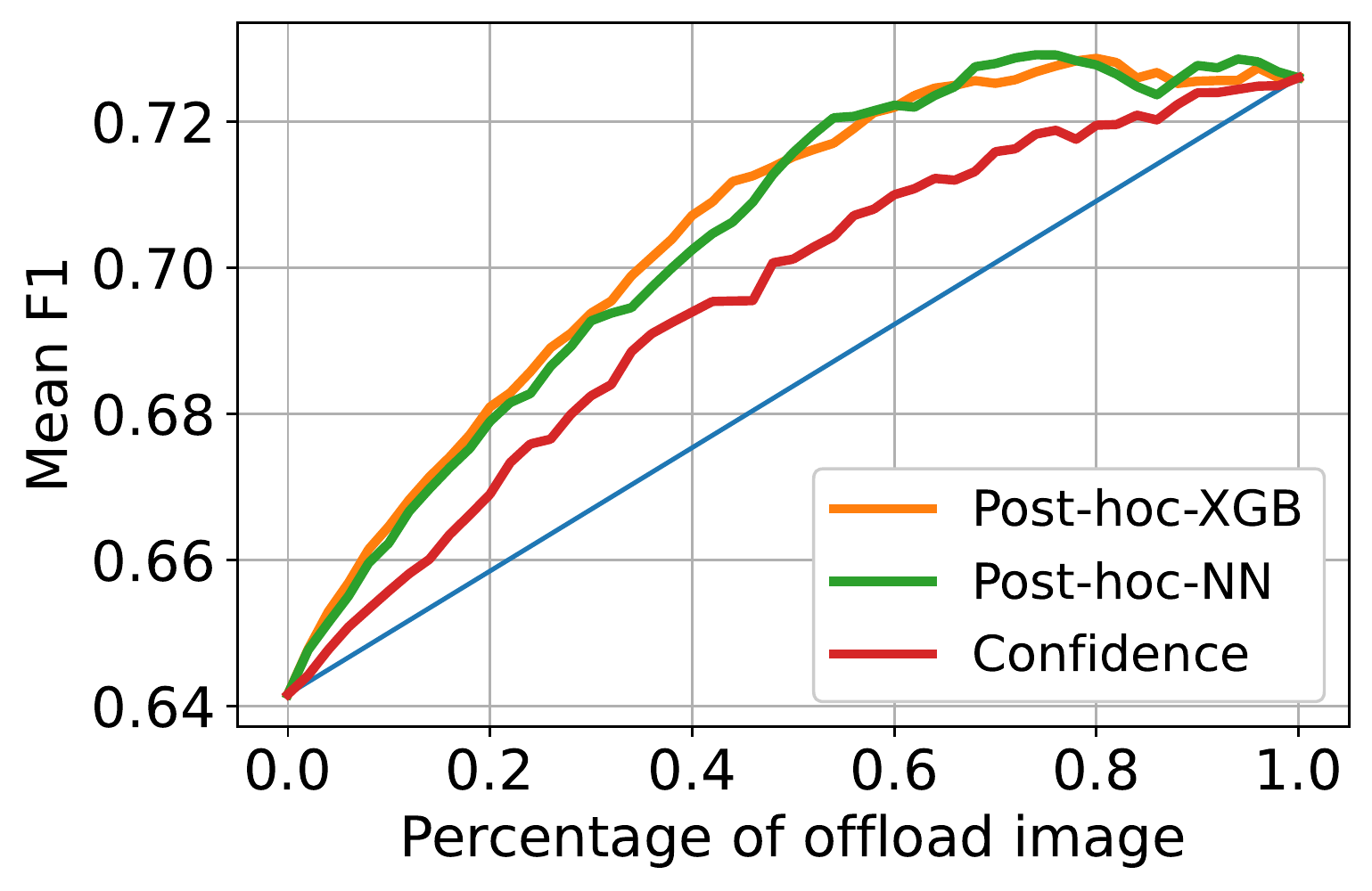}\vspace{-6pt}
			\caption{SSD Inception (client) \& Fast R-CNN ResNet-101 (server)}\label{fig3b}
		\end{subfigure}	
		\caption{Offloading: Comparison of post-hoc models with baselines to decide which samples to offload, for two client-server model pairs. The x-axis is the fraction of images offloaded (as determined by the predicted F1-score difference for the post-hoc models, or confidence for the confidence baseline). 
		}
		\label{fig:offloading}
	\end{figure} 

 Figure~\ref{fig:offloading} plots the mean F1 score of the various methods (ground truth, post-hoc model, and confidence baseline) vs. the fraction of test images offloaded.
 The test images are sorted by the predicted performance difference between the two models (or the confidence, for the confidence baseline).
 A simple thresholding policy is used to determine which images should be offloaded according to their predicted performance difference; a lower threshold leads to more images being offloaded, and a higher threshold leads to more images being offloaded, as discussed in Section \ref{sec:offloading_framework}.

 The results show that as we sweep across thresholds, the offloading fraction increases, more images will be offloaded to the server, and hence the mean F1-score across all images improves.
Our post-hoc model 
(orange/green/red) has a monotonically increasing trend and can increase the mean F1-score across all test images by precisely picking which images to offload.
 The baseline confidence method performs worse as it has lower mean F1-score across offloading fractions; for example, 
 when offloading 20\% of the images (corrresponding to $S_{\text{threshold}}^{\rho}= 0.32$), the improvement of our post-hoc model over the confidence baseline is 0.01 in terms of mean F1 score. 
 When the offloading fraction is 40\% (corresponding to $S_{\text{threshold}}^{\rho} = 0.19$), the improvement is 0.013 (for scale, note that the total average true performance difference is only 0.147).

 \emph{In summary, our post-hoc model accurately predicts the performance difference, resulting in an offloading of certain images to a more powerful black-box model, hence improving the average F1-score across all test images. The average F1 improvement over the confidence baseline is up to 0.013 for SSD Inception + Fast R-CNN ResNet-101.}

\begin{table}
\small
	\centering
	\begin{tabular}{|c|c|}
	\hline
		\textbf{Model} & \textbf{Mean F1-score} \\
		\hline
		SSD MobileNets & 0.46872\\
		\hline
		SSD ResNet-50 & 0.61620\\
		\hline
		SSD Inception & 0.64959\\
		\hline
		Fast R-CNN ResNet-101 & 0.64613\\
		\hline
		Combined (Post-hoc-NN) & 0.65399\\
		\hline
		Combined (Post-hoc-XGB) &0.66841\\
		\hline
		Optimal & 0.70983 \\
		\hline
	\end{tabular}
	\caption{Model selection: Mean F1-score of the individual black-box models, and the model chosen by the post-hoc model.} 
	\label{tab:model_selection}
\end{table}

\subsubsection{Use Case: Model Selection}
\label{sec:model_selection}

How well can the post-poc model predict the performance of several models, and hence choose the right model for a given image?
We first adjusted the classification thresholds of the models to get four models with diverse performance across the different test images (otherwise, one black-box model might always consistently outperform the others).
The post-hoc model is trained as discussed in Section \ref{sec:model_selection}. 
Table~\ref{tab:model_selection} shows the mean F1-score of each model individually (rows 1-4), as well as from our post-hoc model (rows 5-6) that tries to pick the best model.
The results show that 
the post-hoc model can choose the appropriate black-box model and achieve higher F1-score, compared to using a individual model alone.

\begin{figure}
	\centering
		\begin{subfigure}[t]{0.2\textwidth}
		\includegraphics[width=\linewidth]{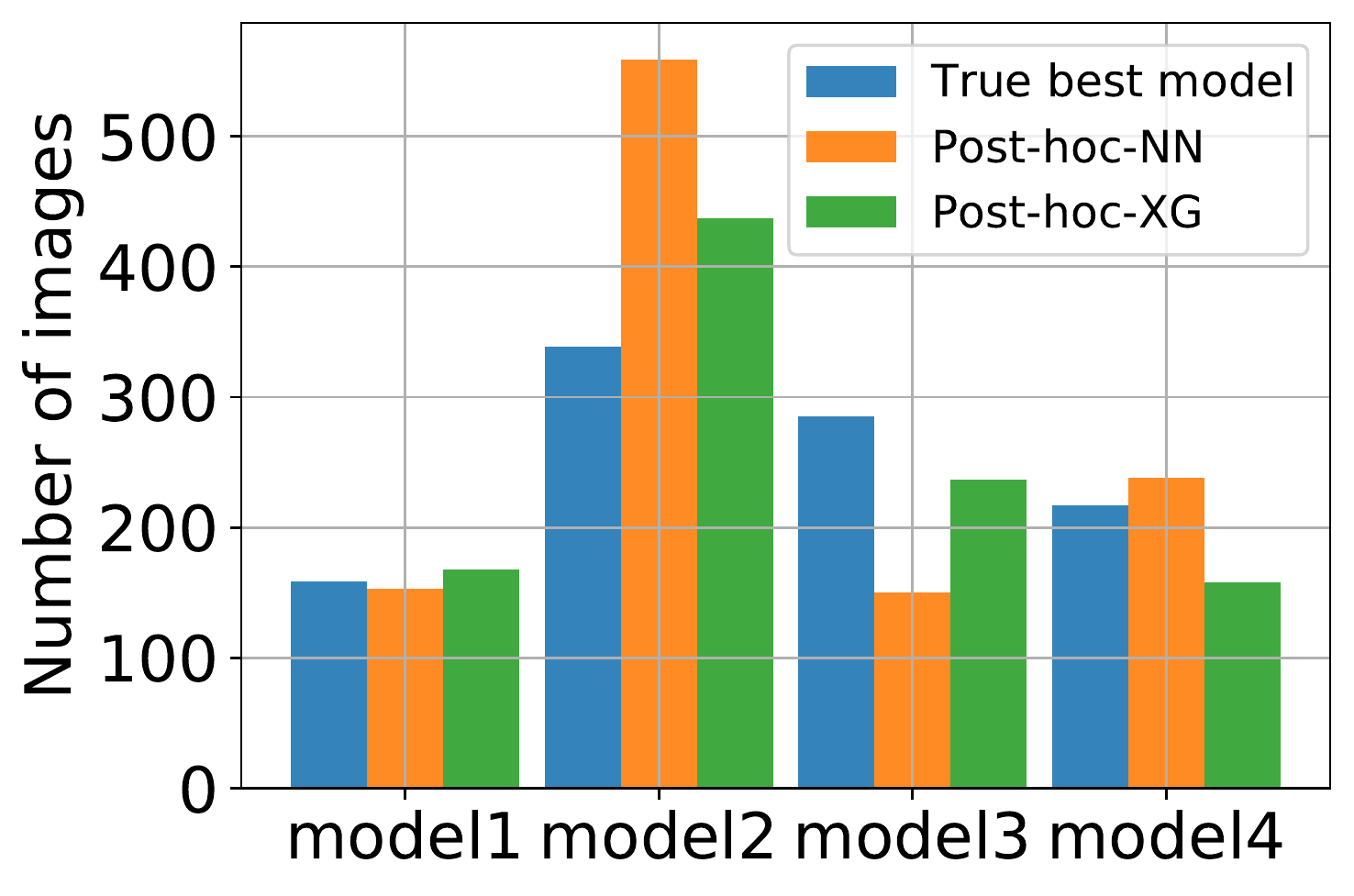}\vspace{-6pt}
		\caption{Histogram of test samples corresponding to each model}\label{fig5a}
	\end{subfigure}
	~
	\begin{subfigure}[t]{0.2\textwidth}
	\includegraphics[width=\linewidth]{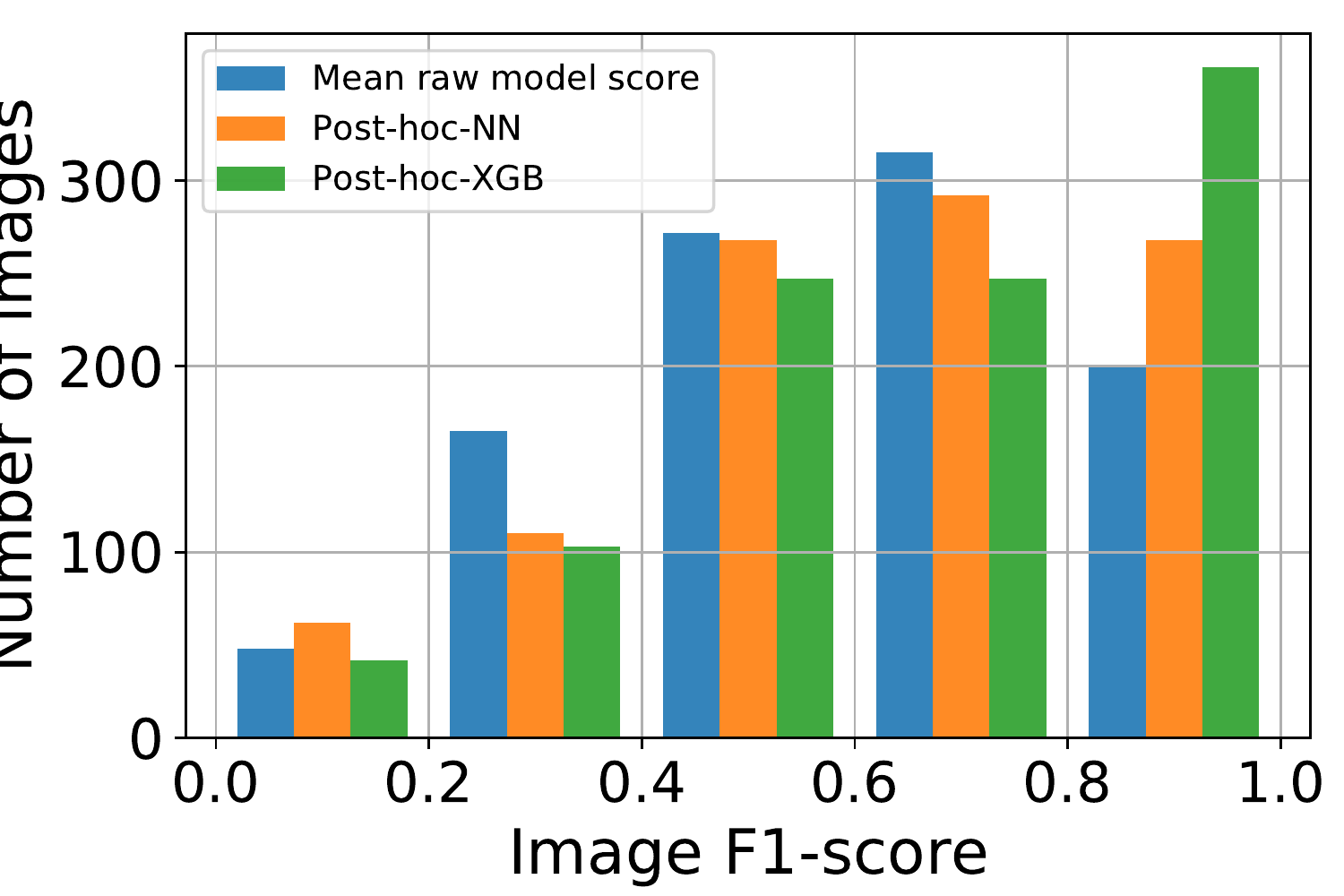}\vspace{-6pt}
	\caption{Histogram of mean F1-score achieved by the different methods}\label{fig5b}
\end{subfigure}
	\caption{Model selection: Distribution of test results. (a) We count the number of images matched to each model by the post-hoc model or the ground truth; more closely matching the true best model (blue) is better. (b) We bin the test samples by F1-score and plot the number of images in each bin; more images with higher F1-score is better. 
	}
	\label{fig:model_selection_decision}
\end{figure} 

To delve deeper into these results, we perform a more detailed analysis of the models chosen by each approach.
Figure~\ref{fig5a} shows the number of test images assigned to each model, according to the optimal policy (labeled as ``true best model''), as well as chosen by the post-hoc model.
Note that although SSD Inception generally has the highest mean F1 score, it's not the optimal choice for all samples.
Ideally, any post-hoc model should match with the true best model (blue bars).
We can see that the distribution of Post-hoc-XGB (orange bars) most closely matches the optimal policy, suggesting that it's able to accurately predict the F1-scores of the images. 

Figure~\ref{fig5b} shows the distribution of F1-score across all the test images, as achieved by the post-hoc models, compared to the average F1-score across all models achieved by each image (labeled ``mean raw model score'' -- essentially, a measure of how easy/difficult the test images are).
Ideally, there will be more samples with high F1-score, and vice versa. In other words, we hope to see the low F1-score bins (left side) contain fewer images, and the high F1-score bins (right side) with more images. 

The results show that the number of images with low F1-score (lower than 0.2) according to Post-hoc-XGB is small, while many images achieve an F1-score between 0.8-1.
\emph{In summary, the post-hoc model is able to select the most suitable black-box model for each image and achieve higher mean F1-score, an improvement of 0.018 over the best-performing solo black-box model.}

\section{Conclusions}
\label{sec:conclusions}

In this paper, we proposed a framework to predict inference performance for a diverse set of metrics and scenarios.
We developed a gradient boosting-based post-hoc model that is easy to train and flexible to use. 
Our model predicts the per-image performance for a variety of metrics, going beyond baseline confidence methods that only predict the class probabilities. By incorporating handcrafted features (\eg the number of bounding boxes) into our post-hoc model, we were able to outperform a confidence baseline.
Future work includes extending the framework to additional performance metrics and use cases, such as hyperparameter optimization for machine learning model architecture search.

\section*{Acknowledments}

This work is supported in part by NSF CAREER awards CCF-2046816 and CNS-1942700, by Army Research Office under MURI Grant W911NF-21-1-0312, and by a Facebook Faculty Research Award.

\ifCLASSOPTIONcaptionsoff
  \newpage
\fi



%

\bibliographystyle{IEEEtran}
\bibliography{Bibfiles,icml}
%








\end{document}